%% file: uai2026-main.tex
\documentclass[accepted]{uai2026/uai2026} 

\usepackage[american]{babel}

\usepackage{natbib} 
\usepackage{mathtools} 
\usepackage{booktabs} 
\usepackage{tikz} 

\usepackage{cleveref}
\usepackage{amssymb}
\usepackage{caption}
\usepackage{subcaption}
\usepackage{float}

\usepackage{color-edits}
\addauthor{kh}{red}
\addauthor{as}{blue}

\usepackage{nicefrac,xspace}

\newcommand{\term}[1]{\ensuremath{\mathtt{#1}}\xspace}

\newcommand{\RawRew}{\term{rew}}
\newcommand{\AveRew}{\ensuremath{\overline{\RawRew}}}

\newcommand{\ie}{i.e.,~\xspace}

\newcommand{\eg}{e.g.,~\xspace}
\newcommand{\Eg}{E.g.,~\xspace}

\newcommand{\rbr}[1]{\left(\,#1\,\right)}

\newcommand{\cbr}[1]{\left\{\,#1\,\right\}}

\newcommand{\eps}{\ensuremath{\epsilon}}

\newcommand{\R} {\ensuremath{\mathbb{R}}} 
\newcommand{\LDOTS}{\, ,\ \ldots\ ,}     

\newcommand{\gptfive}{\textsc{Gpt-5}\xspace}
\newcommand{\gptfivetwo}{\textsc{Gpt-5.2}\xspace}
\newcommand{\gptfivemini}{\textsc{Gpt-5-mini}\xspace}
\newcommand{\gptfivenano}{\textsc{Gpt-5-nano}\xspace}
\newcommand{\gptfour}{\textsc{Gpt-4}\xspace}
\newcommand{\gptfouro}{\textsc{Gpt-4o}\xspace}
\newcommand{\gptthree}{\textsc{Gpt-3.5}\xspace}
\newcommand{\deepseek}{\textsc{DeepSeek-R1-Distill-Qwen}\xspace}
\newcommand{\qwen}{\textsc{Qwen-2.5}\xspace}
\newcommand{\gemma}{\textsc{Gemma-3}\xspace}
\newcommand{\llama}{\textsc{Llama-3-8B}\xspace}
\newcommand{\mistral}{\textsc{Mistral-7B}\xspace}

\newcommand{\FCor}{\ensuremath{\mathrm{FracCorrect}}\xspace}

\newcommand{\xhdr}[1]{\vspace{0.5mm} \noindent{\bf #1}}

\title{Should You Use Your Large Language Model to Explore or Exploit?}

\author[1]{Keegan Harris}
\author[2]{Aleksandrs Slivkins}
\affil[1]{%
    UC Berkeley, \texttt{keegan.harris@berkeley.edu}
}
\affil[2]{%
    Microsoft Research, \texttt{slivkins@microsoft.com}
}
\addtocontents{toc}{\protect\setcounter{tocdepth}{0}}

\begin{document}
\maketitle

\begin{abstract}
    \input{main/abstract}
\end{abstract}

\input{main/intro}
\input{main/related}
\input{main/exploit}
\input{main/explore}

\input{main/conc}
\begin{acknowledgements}
    Some of the results were obtained while KH was a Ph.D. student at Carnegie Mellon University and an intern at Microsoft Research. We would like to thank Dylan Foster, Akshay Krishnamurthy, and anonymous reviewers for helpful comments and suggestions.
\end{acknowledgements}

\bibliography{refs,bib-abbrv,bib-slivkins,bib-AGT,bib-bandits,bib-RL}

\newpage

\onecolumn
\appendix

\section*{Appendix}
\addtocontents{toc}{\protect\setcounter{tocdepth}{2}}
\renewcommand{\contentsname}{}
\tableofcontents
\newpage

\section{Main Plots for the Q/A explore task}
\label[appendix]{sec:QA_app}
\input{app/QA_app}

\clearpage

\section{Details for Section~\ref{sec:exploit}: LLMs as exploitation oracles}
\label[appendix]{sec:exploit_app}
\input{app/exploit_app}

\clearpage

\section{Details for Section~\ref{sec:explore}: LLMs as exploration oracles}\label[appendix]{sec:explore_app}

\input{app/explore_app}

\end{document}

%% file: main/abstract.tex
We evaluate the ability of the current generation of large language models (LLMs) to help a decision-making agent facing an exploration-exploitation tradeoff. 
While previous work has largely study the ability of LLMs to solve combined exploration-exploitation tasks, we take a more systematic approach and use LLMs to explore and exploit \emph{in silos} in various (contextual) bandit tasks. 
We find that reasoning models show the most promise for solving exploitation tasks, although they are still too expensive or too slow to be used in many practical settings. 
Motivated by this, we study tool use and in-context summarization using non-reasoning models. 
We find that these mitigations may be used to substantially improve performance on medium-difficulty tasks, however even then, all LLMs we study perform worse than a simple linear regression, even in non-linear settings. 
On the other hand, we find that LLMs do help at exploring large action spaces with inherent semantics, by suggesting suitable candidates to explore.\looseness-1

%% file: main/intro.tex
\section{Introduction}
The machine learning community is increasingly interested
to apply
advances in generative AI and large language models (LLMs) to
decision-making problems.
Early work in this direction has already produced impressive agentic behavior in both virtual \citep[\eg][]{wang2023voyager, openai2025introducing} and physical-world environments \citep[\eg][]{black2024pi_0}.
Beyond generalization (needed for supervised learning), decision-making under uncertainty requires two additional capabilities:
\emph{exploitation}
(making best decisions given the current data)
and \emph{exploration}
(for long-term benefit).
Balancing the two has led to
a large
literature
\citep[\eg][]{slivkins-MABbook,lattimore2020bandit,RLTheoryBook-20}.

A recent line of work \citep[\eg][]{krishnamurthy2024can, nie2024evolve} evaluates the ability of LLMs to
balance exploration and exploitation entirely \emph{in-context}, \ie  specifying the problem description, parameters, and history in the LLM prompt.
Focused on simple tasks in reinforcement learning (RL), these results are mixed. 
LLMs fail to solve these tasks adequately out-of-the-box, but they can be prompted to do so by providing succinct  summary statistics in-context. However, such statistics do not exist beyond simple
RL problems, \eg for contextual bandits.
RL-specific pre-training or fine-tuning (on data from algorithmic baselines on similar problem instances) tends to work well (see Related Work), but may be prohibitive due to cost or insufficient training data. Besides, using a commonly available
LLM would leverage its ``generalist" intelligence and may be much easier in terms of logistics and required expertise.


Motivated by these observations, we study the ability of LLMs to explore and exploit in-context \emph{in silos}, with an eye towards leveraging a pre-trained LLM (and the inductive bias therein) as a part of a larger decision-making agent.
We focus on (contextual) bandits, as a standard abstraction for the explore-exploit tradeoff.
We experiment with many LLMs: $\gptfivetwo$, $\gptfivemini$, $\gptfivenano$, $\gptfour$, $\gptfouro$, $\gptthree$, $\qwen$, $\gemma$, $\mistral$, and $\deepseek$, a reasoning model.
\footnote{Some other LLMs did not work for our purposes: \textsc{Llama-3.2-3B}  
would not follow instructions. Two reasoning models, \textsc{DeepSeek-R1-Distill-Llama-70B} and \textsc{Qwen-2.5-Math-70B}
have a context window that is too short.}\looseness-1

In~\Cref{sec:exploit}, we evaluate LLMs as \emph{exploitation oracles} for contextual bandits.
%
Given a history of (context, action, reward) tuples, the LLM is tasked with identifying the best action to take given a new context.
Our results here are mixed. We show that LLMs can effectively exploit in-context for small-sized
tasks
but their performance degrades when the
tasks
become moderately sized.
%
We find that frontier reasoning models outperform non-reasoning models
(fixing model size and provider),
but tend to be slow and/or expensive.
%
We investigate mitigations: in-context summary techniques and tool use
(namely, a Python code interpreter).
We find that both
mitigations improve
performance in a linear task, but on non-linear tasks LLMs with these mitigations still perform worse than a simple linear regression baseline.\looseness-1

In~\Cref{sec:explore}, we evaluate LLMs as an \emph{exploration oracle} which suggests a small set of candidate actions by discretizing a large action space. (This set can then be used to instantiate an off-the-shelf bandit algorithm.) On various text-based multi-armed bandit tasks that we design, we find this approach far superior to a non-LLM baseline.
%
Our first task is based on the MovieLens dataset~\citep{harper2015movielens}, where actions
are movie recommendations, rewards are random draws parameterized by ground-truth movie ratings,
%
and the baseline is
a hand-crafted discretization based on movie genre.
%
Second, we design
a larger-scale bandit task based on paper titles/abstracts from arXiv:
find a suitable title for a given abstract.
Given the high dimensionality of the action space,
discretization approaches
from
continuous bandit problems are inapplicable.
%
Several prompting strategies
all lead to relatively good exploration compared to natural baselines. Our third task is based on open-ended “philosophical" questions and contrarian answers. 

%% file: main/related.tex

\xhdr{Related work.}
Our results belong to a 
growing line of work on using pre-trained LLMs for in-context reinforcement learning (RL).
\citet{coda2023meta, krishnamurthy2024can, nie2024evolve, monea2024llms, xia2024beyond, park2024llm, wu2024smartplay} evaluate the ability of LLMs to solve various multi-armed bandit and contextual bandit tasks, and find that the current generation of LLMs largely fail to solve these tasks in-context.
Indeed, positive findings are restricted to very simple tasks and/or require substantial mitigations (which in turn do not readily extend beyond simple settings).
\citet{xia2024beyond} use LLMs to solve dueling bandit tasks, and~\citet{park2024llm} also evaluate the ability of LLMs to learn in games.
While our paper is primarily concerned with whether LLMs succeed as algorithms, several others
\citep[\eg][]{schubert2024context,hayes2024relative,coda2024cogbench} use in-context bandits (and many other tasks) to study whether LLMs exhibit human-like behavior/biases in decision-making.

A broader literature on in-context learning (starting from~\citet{brown2020language}) aims to solve various tasks by providing all relevant information in the LLM prompt.
The work on \emph{exemplar selection} (selecting examples and other information to present in-context)
\citep[\eg][]{khalifa2023exploring, zhang2022automatic, xiong2023dq, tonglet2023seer} is relevant to our exploitation experiments.

A growing line of work aims to use LLMs as a part of a larger decision-making agent \citep[\eg][]{li2024stride, zhou2023language, zhao2024large, harris2026context}.
Our exploration experiments take inspiration from the work on using LLMs as ``action priors'' inside of a larger RL algorithm~\citep{yan2024efficient, carta2023grounding, yao2024tree, hao2023reasoning}.
Much of this work falls under the proposer-verifier framework of~\citet{snell2024scaling}, where an LLM proposes several possible sequences from which a verifier selects suitable candidates.
In comparison, our goal is a more systematic evaluation of LLMs' abilities to explore large action spaces, in isolation from other components of the decision-making task.\looseness-1

Finally, a parallel line of work trains transformers to solve various RL tasks \citep[\eg][]{laskin2022context, lin2023transformers, lee2024supervised, raparthy2023generalization, xu2022prompting, lehnert2024beyond, mukherjee2024pretraining}.
Our negative results in~\Cref{sec:exploit} provide an additional foundation for this work, emphasizing the shortcomings on frontier LLMs that are not custom-trained for RL.\looseness-1

\xhdr{Background on multi-armed bandits (MAB).}
%
We consider tasks based on MAB and contextual bandits (CB), well-studied special cases of RL that abstract explore-exploit tradeoff 
\citep{slivkins-MABbook,lattimore2020bandit}.
%
In MAB, there are $T$ rounds and $K$ arms. In each round $t \in [T]$, the learner chooses an action (\emph{arm}) $a_t \in [K]$ and observes reward $r_t$ drawn from some (fixed, unknown) sub-Gaussian reward distribution $D(a_t)$
with mean $\mu(a_t)$.
%
%
In CB, the learner also
observes a context $z_t$ before each round $t$. The expected reward $\mu(z_t, a_t)$ depends on both the context and the arm.
The learner's goal is to balance exploration and exploitation to maximize cumulative reward.

An ``exploitation oracle" (which optimizes for the current round given the history) naturally plugs into standard bandit algorithms such as Epsilon-Greedy, Explore-then-Commit, and Follow-The-Perturbed-Leader. Typical implementations in CB involve model-based (\eg linear) regression or cost-sensitive classification \citep[][Ch.8]{slivkins-MABbook}. Designing CB exploitation oracles for their own sake, a.k.a. \emph{offline policy optimization}, is well-studied (starting from, \eg \citet{beygelzimer2009offset,DR-StatScience14}). 
Usage of an LLM as an ``exploration oracle" is closely related to the literature on dynamic pricing and Lipschitz bandits
\citep[\eg][]{KleinbergL03,LipschitzMAB-stoc08,LipschitzMAB-JACM,xbandits-nips08},
as we elaborate in \Cref{sec:explore}.

%% file: main/exploit.tex
\section{LLMs as exploitation oracles}\label{sec:exploit}

\begin{figure*}[t]
    \centering
    \includegraphics[width=\linewidth]{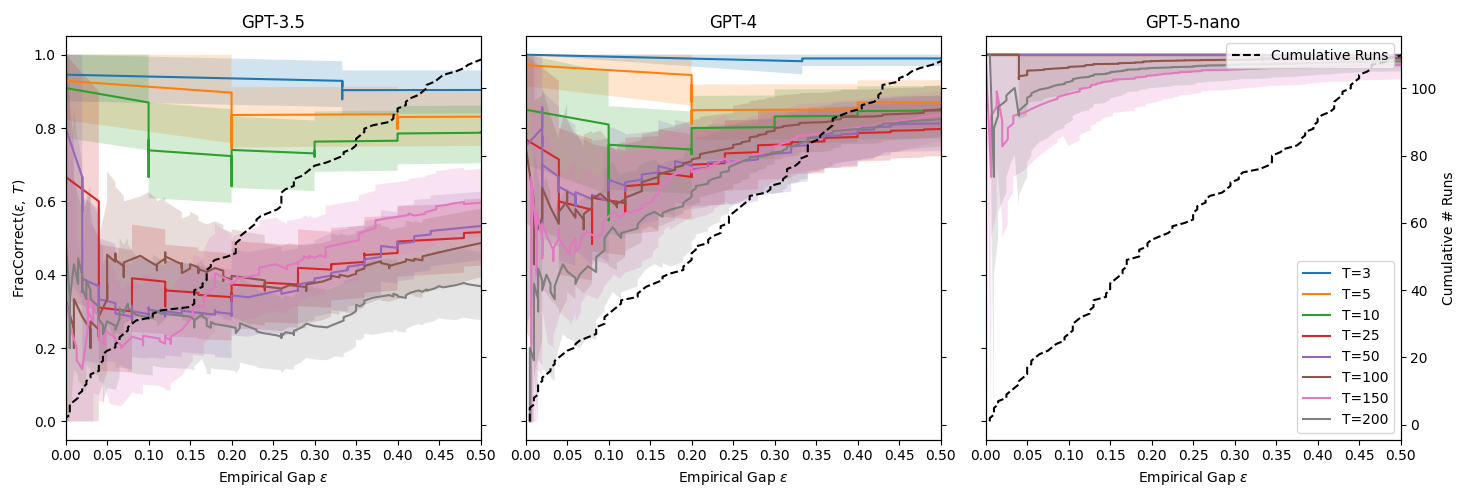}
    \vspace{-5mm}
    \caption{MAB exploit puzzle for  $\gptthree$ (left), $\gptfour$ (middle), and $\gptfivenano$ (right), all with the ``buttons'' prompt and temperature $1$. Results for $\gptfivemini$ are not visualized, as $\FCor(\eps,T) = 1$ for all $\eps$ and $T$. The following conventions apply to all figures in this section. Each line corresponds to a particular value of \#rounds $T$ and plots $\FCor(\eps,T)$ against empirical gap $\eps$ on the X-axis. The shaded band around the line represents a $95\%$ confidence interval. The dashed line is the number of tasks (``runs") with empirical gap $\leq\eps$; the resp. Y-scale is on the right.\looseness-1 }
    \label{fig:mab-main}
\end{figure*}

We evaluate the ability of LLMs to \emph{exploit} in decision-making tasks with statistical uncertainty on the outcomes. We present LLMs with in-context exploit tasks inspired by multi-armed bandits (MAB) and contextual bandits (CB). In a CB exploit task, an LLM is given a history consisting of context-arm-reward tuples, and is instructed to take the best arm given the current history and the current context. A MAB exploit task is the same, but without contexts. These tasks are generated from some parameterized distributions called \emph{exploit puzzles}.
Due to computational constraints, our experiments focus mostly on
$\textsc{Gpt}$ model family,
with additional LLMs evaluated as a robustness check in Appendix~\ref{sec:exploit-robustness}, with similar findings.
Across all
experiments,
varying the model temperature did not significantly affect performance.
%
We include results for temperature 0 and 1.




\xhdr{MAB exploit puzzles.}
We begin by
testing the $\textsc{Gpt}$ model family
on simple MAB exploit puzzles.
We find that performance significantly improves as the models progress
from $\gptthree$ to $\gptfour$
to $\textsc{Gpt-5}$.
In fact, $\gptfivemini$ achieves perfect performance.
%
Our experiments on $\gptfour$ and $\gptthree$
provide a partial explanation for the observation
in \citep{krishnamurthy2024can, nie2024evolve},
that these models fail to solve end-to-end MAB tasks in-context
when presented with raw (non-summarized) history.\looseness-1

Following these two papers, we try two prompts, where arms correspond, resp., to pushing different colored buttons and to showing different ads to users.
The LLM is asked to choose the arm with the highest empirical reward in the next round.
We also try chain-of-thought (CoT) prompts, for the total of 4 prompt designs: $\cbr{\text{buttons, adverts}} \times \cbr{\text{CoT, no-CoT}}$.
See \Cref{app:prompts} for more details on this
setup.

Our MAB exploit puzzle is parametrized by gap $\Delta\in [0,1]$ and history size $T$. The tasks, all with 5 arms, are constructed as follows. We pick an arm $a^*$ uniformly-at-random (u.a.r.). Expected rewards are
$\mu(a^*) = \nicefrac12 + \nicefrac{\Delta}{2}$
and
$\mu(a) = \nicefrac12 - \nicefrac{\Delta}{2}$
for all other arms $a$. Then, we generate a history of $T$ rounds for each arm $a$, where the reward $r_t(a)$ at each round $t\in[T]$ is an independent Bernoulli draw with mean $\mu(a)$.
For a given $T$, we generate 10 tasks from this puzzle for each value of
$\Delta \in \{0,\, .05,\, .1,\, .15,\, .2,\, .25,\, .3,\, .4,\, .45,\, .5\}$.\looseness-1

\begin{figure*}[t]
     \centering
     \begin{subfigure}[b]{0.49\textwidth}
        \centering
        \includegraphics[width=\linewidth]
        {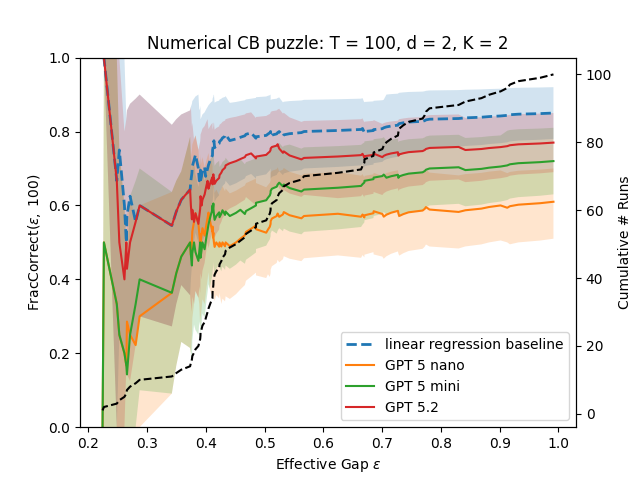}
     \end{subfigure}
     \hfill
     \begin{subfigure}[b]{0.49\textwidth}
        \centering
        \includegraphics[width=\linewidth]
        {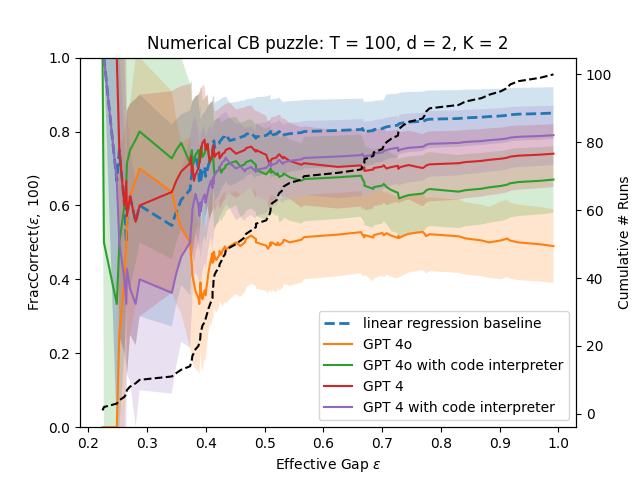}
     \end{subfigure}
     \caption{%
     $\textsc{Gpt-5}$ (left) and \gptfour (right) model families on a numerical CB puzzle (all with temperature $1$). $\textsc{Gpt-5}$ models generally perform better than \gptfour models of similar size, although letting $\gptfour$ and $\gptfouro$ use a code interpreter improves performance.\looseness-1}
     \label{fig:numerical}
\end{figure*}

Given a (realized) exploit task, we measure its difficulty via \emph{empirical gap} $\eps$: the difference between the largest and second-largest average reward
    $\bar{r}_t(a) :=  \frac{1}{T} \sum_{t\in [T]} r_t(a)$
among all arms $a$. Intuitively, puzzle difficulty decreases with $\eps$ (as it gets harder to distinguish the top two arms). Empirical gap mirrors the ``gap" between the top two expected rewards, a standard notion of difficulty in MAB.

We measure an LLM's performance over a given set $S$ of tasks as the fraction of tasks for which the LLM returns a ``correct answer'': an arm with the largest empirical reward; denoted by $\FCor(S)$. We are interested in how \FCor varies depending on the difficulty level. Hence, let $S(\eps,T)$ be the set of all tasks with empirical gap at most $\eps$ and history size $T$.
We plot
$\FCor(\eps,T) := \FCor\rbr{S(\eps,T)} $
against
$\eps$.

We find that while $\gptfour$ and $\gptthree$ do not perform well,
$\textsc{Gpt-5}$ models do, see \Cref{fig:mab-main}.
For $\gptfour$ and $\gptthree$, performance tends to degrade as (1) history size  $T$ increases, with more data points to process in-context and (2) empirical gap decreases, making the bandit instance more difficult.
%

%



\xhdr{CB exploit puzzles.}
While the history in $K$-armed bandits can be summarized with $2K$ numbers (for each arm, the average reward and \#plays),
such succinct summary statistics may not be readily available (or even exist) in more complicated decision-making tasks such as CB.%
\footnote{This consideration also motivates MAB exploit puzzles with raw (non-summarized) history, as a simpler special case of the general scenario when succinct summarization is unavailable.}\looseness-1


We begin with linear CB, where
the expected reward of each arm $a$ is linear in the context $z\in \R^d$:
$\mu(z,a) = \langle z, \theta^*_a \rangle$ for some fixed (but unknown) vector $\theta^*_a\in\R^d$.
\footnote{Linear CB are well-studied, starting from \citet{Langford-www10,Reyzin-aistats11-linear,Csaba-nips11}.}
(We turn
to non-linear CB later in the section.)

We consider a CB exploit puzzle parameterized by \#arms $K$, dimension $d$, and history size $T$.
The tasks are constructed as follows. We sample parameters $\theta_a \in [-1, 1]^d$ and $\gamma_a \in [-\frac14,\frac14]$ independently u.a.r. for each arm $a$. Given context $z\in \mathbb{R}^d$, expected reward for arm $a$ is $\mu(z, a) = \langle z, \theta_a \rangle + \gamma_a$. We generate a history of $T$ rounds $t\in [T]$. Contexts $z_t$ are sampled independently u.a.r. from $[-1, 1]^d$. For simplicity, the history contains rewards of all arms $a$ in all rounds $t$, where the reward equals $ \mu(z_t,a)$ plus an independent unit-variance Gaussian. Given the history and a new context $z_{T+1}$ (drawn in the same way), the LLM is asked to select the action for round $T+1$ which appears best. This gives one exploit task. We generate $N$ tasks for the same $K,T,d$.\looseness-1

We use a modified ``buttons'' prompt, where contexts correspond to ``numbers on a screen" which affect the payoffs for each button. The prompt does not mention linearity of the CB instance (because such model-based information is typically unavailable in applications).


Given an exploit task, we now define a ``correct answer" as arm $a$ which maximizes expected reward $\mu(z_{T+1},a)$.%
\footnote{Note that it is unclear how to define an "empirically best arm" given a CB history and the current context.} $\FCor(S)$ is the fraction of correct answers in a given set $S$ of tasks.
%
%
%
Task difficulty is also not easily defined in terms of realized rewards. Instead, we focus on \emph{effective gap} $\eps$: the difference in expected reward $\mu(a,z)$ between the best and second-best arm given the current context $z = z_{T+1}$.
Intuitively, smaller $\eps$
corresponds to increased difficulty.

We study
how $\FCor$ varies with effective gap. In each plot,
we let $S(\eps,T)$ be the set of all tasks with given $K,d,T$ and effective gap at most $\eps$.
We plot
    $\FCor(\eps,T):= \FCor\rbr{S(\eps,T)}$
against $\eps$.

\begin{figure*}[t]
     \centering
     \begin{subfigure}[b]{0.49\textwidth}
        \centering
        \includegraphics[width=\linewidth]{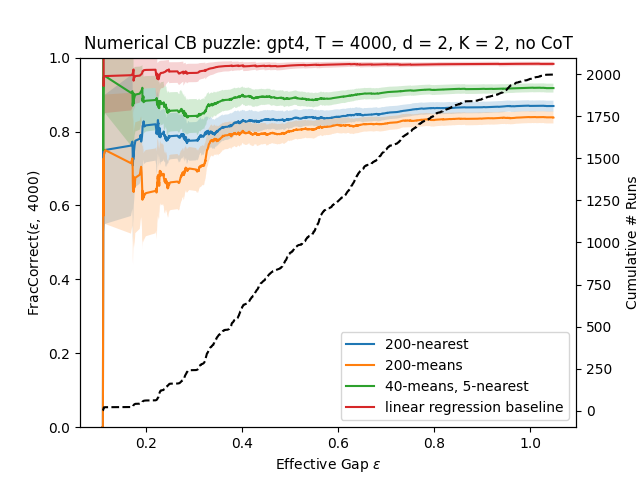}
     \end{subfigure}
     \hfill
     \begin{subfigure}[b]{0.49\textwidth}
        \centering
        \includegraphics[width=\linewidth]{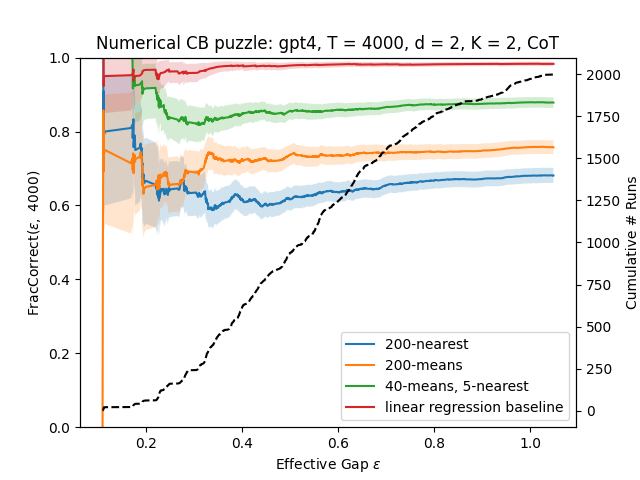}
     \end{subfigure}
    \caption{CB exploit puzzle with $d=K=2$, $T=4000$, and temperature $0$: \textbf{mitigations help substantially}. $\gptfour$ without CoT (left) and $\gptfour$ with CoT (right). 
    (Full history with this $T$ vastly exceeds the context window for $\gptfour$, $\gptfouro$, and $\gptthree$.)}

    \label{fig:cb-main1}
\end{figure*}

\Cref{fig:numerical} considers a setting with
$d = K = 2$ for the $\gptfive$ family
(left) and
the $\gptfour$ family
(right).
We find that larger models generally perform better than their smaller, distilled counterparts and, conditioned on model size, $\gptfive$ models outperform $\gptfour$ models.
We also plot a linear regression baseline (blue dashed line), which effectively serves as an upper-bound on the performance of any model, due to the underlying linear structure of our
CB puzzle.

\begin{figure*}[t]
    \centering
    \includegraphics[width=\linewidth]{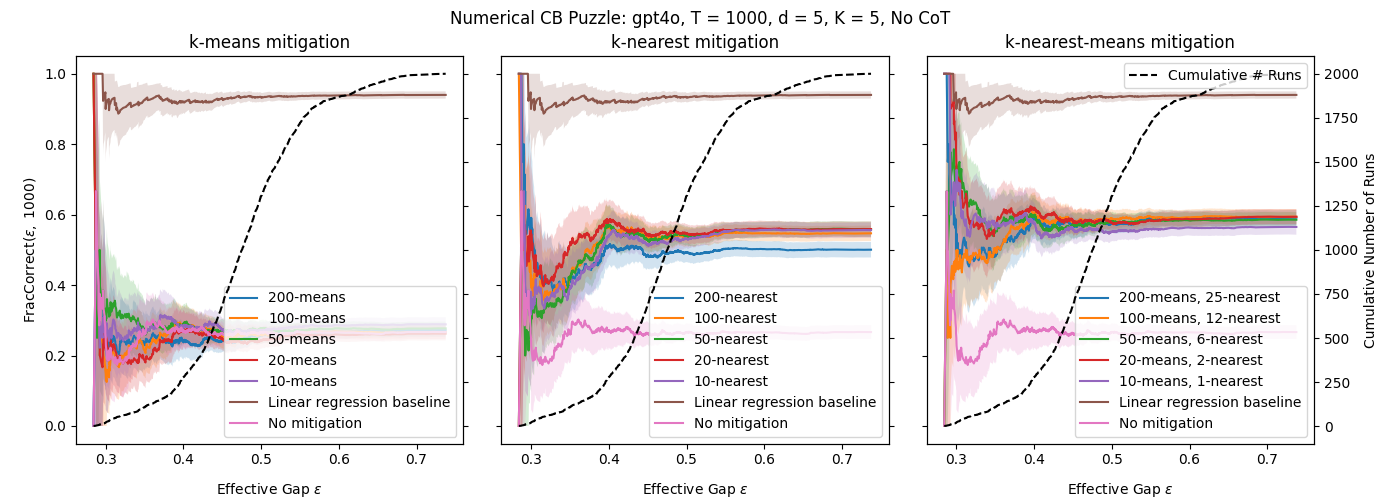}
 \caption{CB exploit puzzle with $d=K=5$, $T=1000$, and temperature $0$: \textbf{in-context mitigations perform badly}, but (mostly) much better than the no-mitigation baseline. $\gptfouro$ without CoT.}
    \label{fig:cb-main2}
\end{figure*}

While not pictured, all models tend to perform strictly better on smaller problem instances.
For example, $\gptfour$
performed near-perfectly when
$d=1$, $K=2$, and $T=50$.
However, performance
degrades as the problem size grows (e.g., see $\gptfouro$ in~\Cref{fig:cb-main2}).
%
Moreover, limited prompt size may prevent processing larger histories.\footnote{\Eg our LLM access points bottomed out at $T\approx \text{100-200}$ for $\gptfour$ and $T\approx \text{1000-2000}$ for $\gptfouro$.}\looseness-1

Motivated by these observations, we implement two types of \emph{mitigations}: tool use (providing the model with access to a Python interpreter to execute code in) and in-context summarization (inspired by the literature on exemplar selection for in-context learning, see Related Work).
Evaluating these mitigations, we focus on the $\gptfour$ family as (1) at the time of writing $\gptfive$ models are not configured for tool use and (2) we found the $\gptfive$ models to be either prohibitively expensive
(%
$\gptfivetwo$) or prohibitively slow
(%
$\gptfivemini$ and $\gptfivenano$) on larger-scale experiments.\looseness-1

See~\Cref{fig:numerical} (right) for the performance of $\gptfour$ and $\gptfouro$ with access to a code interpreter (CI).
%
Both see significantly improved
performance with CI,
with $\gptfour$ even surpassing
$\gptfivetwo$.
However, enabling the CI made
$\gptfour$ and $\gptfouro$ much slower and much more expensive.%
\footnote{E.g., our $\gptfour$ experiment with CI
in~\Cref{fig:numerical} cost approximately $\$70$ and took
6+ hours to run.
Our $\gptfour$ experiment in~\Cref{fig:numerical} without CI
cost a few dollars and took under 5
minutes.}
\looseness-1

We consider the following summarization mitigations:

 \noindent \textbf{1.} \emph{$k$-nearest:}
 Among the observed contexts,
 pick
 distinct $k$ contexts closest to $z_{T+1}$, according to the $\ell_2$ metric.
 Limit the
 reported history
 to (the rounds with) these $k$ contexts.

 \noindent \textbf{2.} \emph{$k$-means:} Run an off-the-shelf algorithm for $k$-means clustering on contexts $\cbr{z_1 \LDOTS z_T}$, obtaining $k$ centroid contexts $z^*_i$ and their respective clusters $Z^*_i$, $i\in [k]$. For each centroid $z^*_i$ and each arm $a$, let $\bar{r}(z^*_i,a)$ be the average reward  for this arm over all rounds $t$ with contexts $z_t\in Z^*_i$. Report $\rbr{z^*_i,\,a,\,\bar{r}(z^*_i,a)}$ as a context-arm-reward triple.

 \noindent \textbf{3.} \emph{$k$-means, $k'<k$-nearest:} First, run the $k$-means mitigation. Report $\rbr{z^*,\,a,\,\bar{r}(z^*,a)}$ as a context-arm-reward triple, for each arm $a$ and each centeroid context $z^*$ among the $k'$ centroids closest to $z_{T+1}$ (according to the $\ell_2$ metric).

We do not explain the ``mitigation strategy" in the prompt: we present the context-arm-reward tuples as if it were the entire history without mentioning clustering or averaging.\looseness-1


\Cref{fig:cb-main1} shows the performance of our mitigations on a slightly larger
puzzle with $K=d=2$, $T=4000$. We use $\gptfour$ with and without chain-of-thought (CoT) prompting. As before, we compare
against a linear regression baseline.
%
Without CoT,
all three mitigations achieve \FCor $\approx 80\%-90\%$, although this dips to $\approx 60\%-85\%$ with CoT.
%
Performance aside, the mitigations offer a practical way to handle large histories; e.g., when $T=4,000$, our prompt exceeds the context window of all models we had access to.\looseness-1


However,
in-context mitigations can only help so much.
In~\Cref{fig:cb-main2}, we plot their performance
under various hyperparameters, on $\gptfouro$.
%
While $k$-means (left) is almost as bad as random guessing,
$k$-nearest
(center) achieves $\approx 50\%$ \FCor, and the $k$-means, $k'$-nearest mitigation (right) obtains $\approx 60\%$.
%
The latter two
out-perform unmitigated $\gptfouro$, but fall significantly short of the linear baseline.
%



\xhdr{CB exploit puzzles (text-based \& non-linear).}
As a robustness check, we repeat our CB experiments on a text-based exploit puzzle.
In this puzzle, contexts are items in a room (e.g. animals, objects on a table), and actions have an associated semantic meaning (e.g. eat the food item, leave the room).
Rewards are still presented numerically, and are non-linear functions of both the context and action.
See \Cref{sec:exploit_app} for full details on our experimental setup.

\begin{figure*}[t]
     \centering
     \begin{subfigure}[b]{0.49\textwidth}
        \centering
        \includegraphics[width=\linewidth]{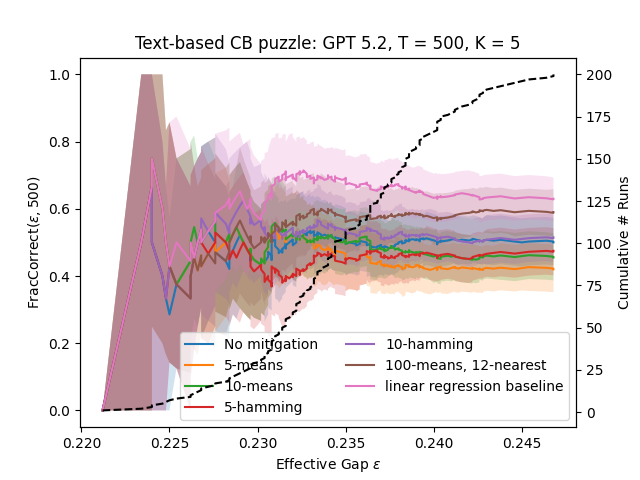}
     \end{subfigure}
     \hfill
     \begin{subfigure}[b]{0.49\textwidth}
        \centering
        \includegraphics[width=\linewidth]{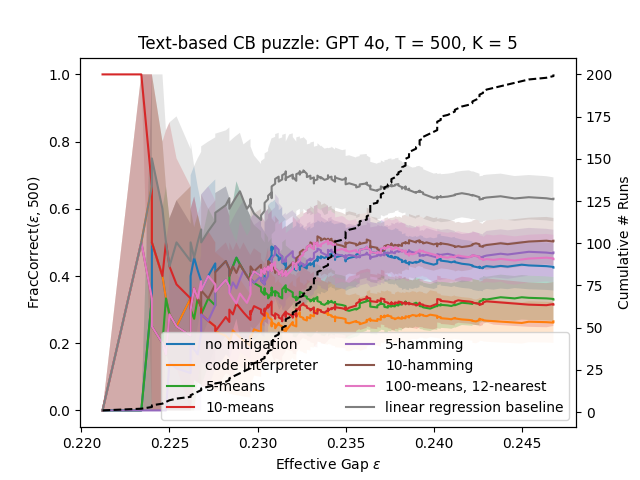}
     \end{subfigure}
    \caption{Performance of the $\gptfivetwo$ (left) and $\gptfouro$ on a text-based CB puzzle with temperature $1$. In this more challenging non-linear setting, some mitigations help, but all are \textbf{outperformed by linear regression}.}
    \label{fig:words}
\end{figure*}

\Cref{fig:words} shows the performance of $\gptfivetwo$ (left) and $\gptfouro$ (right) with mitigations on this puzzle.
While the reward function is non-linear (and thus the linear baseline only achieves ~$70\%$ \FCor), we find that all configurations are still significantly out-performed by the linear baseline.
Interestingly, $\gptfouro$ with code interpreter access (orange) performs the worst out of all configurations.\looseness-1

Our intuition for these findings is as follows: While LLMs are generally good with textual inputs, the reward data is numerical, so the LLM faces same challenges as before  (choosing a ``solution strategy'' and executing it correctly), exacerbated by non-linearity. Meanwhile, linear regression is known to often work fairly well even on non-linear data.

Finally, as a robustness check, we implemented our CB experiments using the CB setting in BanditBench~\citep{nie2024evolve}, which itself is based on the MovieLens dataset \citep{harper2015movielens}.
Here, contexts contain information about users, and the goal is to recommend the best movie out of 20 possible choices to a each new user.
Our findings carry over: we find that the $k$-nearest and $k$-nearest-means mitigations do help, but are still outperformed by the linear regression baseline.
See \Cref{app:bb} for details.

\xhdr{Takeaways.}
Frontier models can now
solve fairly complicated MAB exploit puzzles.
In contrast, $\gptfour$ and $\gptthree$ struggle,
partially explaining their failure on
end-to-end MAB tasks in~\cite{krishnamurthy2024can, nie2024evolve}.

In numerical CB puzzles, unmitigated performance drops as $d,K,T$ grow, but tool use can help and increases in $T$ can largely be taken care of by our in-context mitigations.
In contrast, our mitigations were largely unhelpful in a non-linear, text-based CB puzzle.
Taken together, our findings suggest that while current LLMs are effective at exploiting in small-scale, numerical decision-making tasks, they still struggle in more complicated settings.\footnote{It is worth noting that our results on exploit puzzles can be easily restated in terms of regret for the explore-then-commit algorithm on the corresponding bandit instance. In particular, each plot in Figures \ref{fig:mab-main}-\ref{fig:words} can have a ``twin'' where the Y-axis is regret averaged over the same collection of problem instances (i.e., all instances with effective gap $\leq \epsilon$).}


%% file: main/explore.tex
\section{LLMs as exploration oracles}\label{sec:explore}

We now study the ability of LLMs to explore large action spaces.
We leverage the inductive bias of an LLM to generate a small set of candidate actions from a text-based action space, before running an off-the-shelf MAB algorithm on this set. We refer to this LLM usage as \emph{exploration oracle}.
%

This usage of LLMs
is closely aligned with
``discretization" in
dynamic pricing and Lipschitz bandits~\citep[\eg][]{KleinbergL03,LipschitzMAB-stoc08,LipschitzMAB-JACM,xbandits-nips08,contextualMAB-colt11}.
%
Given a very large action space, it may help
to focus on a much smaller set of candidate arms.
A naively constructed discretization, e.g., uniform
w.r.t. some known distance notion
on the arms,
may have a poor tradeoff between size (\#candidates) and quality (of the best candidate), placing all but a few candidates in low-performing regions of the action space.
A crucial theme in this literature is ``smarter'' discretization algorithms which gradually ``zoom in" on better-performing
arms. If the latter
comprise a low-dimensional region of the
action space, covering this region should require
much fewer samples for the same ``discretization quality''.
Our hope is that LLMs can suggest candidate actions from this region, if there is sufficient semantic meaning encoded in the bandit task.

Non-LLM approaches appear hopeless for our tasks. Standard bandit algorithms (Thompson Sampling, UCB1, etc.)  fail for a very large \#arms, and so do bandit algorithms which randomly subsample the arms. Another approach embeds the arms' text labels in $\mathbb{R}^d$. If a suitable reward-distance relation were known to the algorithm/agent (assumption we do not make), one could apply the ``smarter'' discretization algorithms mentioned above. However, their regret scales exponentially in $d$, making them impractical for state-of-art text embeddings (e.g., one we use has $d=384$).

\xhdr{Our tasks.}
We consider stylized exploration tasks of the following form. In each task, we define an MAB instance with Bernoulli rewards and a very large \#arms, each labeled with a text string. The mean rewards $\mu(\cdot)$ and the meaning of the arms' labels are based on real-world data. We call an LLM once to generate a small representative subset of $K$ arms. To grade the entire set (not just the best arm therein), we run a standard bandit algorithm (UCB1; \cite{bandits-ucb1}) over these $K$ arms, for $T=1000$ rounds. We record the algorithm's average cumulative reward, ``denoised" as
    $\RawRew:= \tfrac{1}{T}\,\sum_{t\in[T]} \mu(a_t)$.

We consider
three task types: (1) recommend a movie,
based on the MovieLens 100K dataset~\citep{harper2015movielens}, a standard benchmark for evaluating CB (MovieLens task), (2) suggest a title for an arXiv research paper given its abstract (arXiv task), and (3) answer an open-ended ``philosophical" question (Q/A task).
Particular workloads within each task type are called  \emph{explore puzzles}.
Our findings are similar across all task types and all LLMs.

The MovieLens 100K dataset
contains ratings from 943 users across 1,682 movies.
In our tasks, arms correspond to the movies titles. Each task corresponds to a particular user, with mean rewards given by this user's ratings.
Since every user did not rate every movie, we used a standard algorithm to complete the data matrix:
\texttt{soft-impute} from the \texttt{fancyimpute} Python package.%
\footnote{Replacing \texttt{soft-impute} with Truncated SVD for data imputation yields similar results, see \Cref{app:SVD-imputation}.}
 We then normalize this matrix so that all values lie $[0,1]$.\looseness-1

The arXiv task (resp. Q/A task) is constructed as follows.
Each task corresponds to an abstract-title pair from arXiv (resp., a question-answer pair). The arms are all possible titles (resp., answers).
The ``best arm" $a^*$ is the actual title (resp. a contrarian answer generated by another LLM).
The expected reward $\mu(a)$ of an arm $a$ is the cosine similarity between $a$ and $a^*$
in the embedding
generated by the Sentence-BERT embedding model~\cite{reimers2019sentence}.
\footnote{While cosine similarity ranges on $[-1,1]$, it was usually strictly positive in our experiments. In the (very rare) cases where it was negative we defined the expected reward as zero.}
\footnote{Our results are largely unchanged under the Universal Sentence Encoder~\cite{cer2018universal}, see 
\Cref{sec:another-embedding-model}.}

As a robustness check, we try several prompting strategies (detailed in \Cref{app:explore-prompts}).
We either ask the LLM to generate suggestions ``all-at-once'' with temperature $0$ or ``1-by-1'' with temperature $1$ (we repeatedly show the LLM the  candidate arms so far and ask it to generate one more).
We also try to explicitly encourage the LLM to provide a diverse candidate set. Thus, we have 4 prompting strategies: $\cbr{\text{all-at-once, 1-by-1}} \times \cbr{\text{with, without}}$ encouragement. However, all 4 strategies performed similarly.%
\footnote{We only tried the 4 strategies for the arXiv and Q/A tasks.}


%


\begin{figure*}[t]
     \centering
     \begin{subfigure}[b]{0.49\textwidth}
        \centering
        \includegraphics[width=\linewidth]{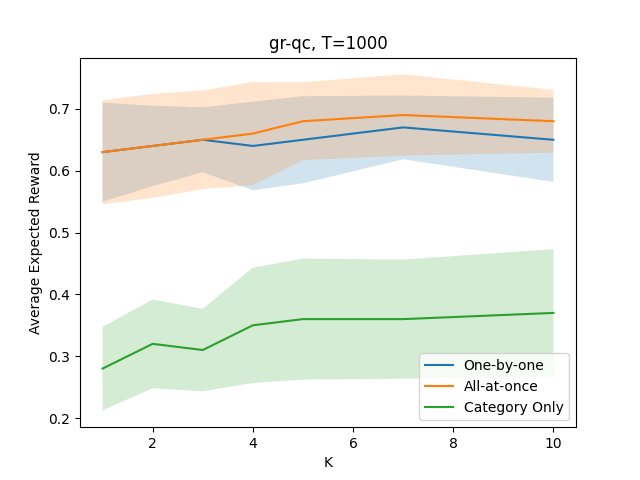}
        \label{fig:explore_gr}
     \end{subfigure}
     \hfill
     \begin{subfigure}[b]{0.49\textwidth}
        \centering
        \includegraphics[width=\linewidth]{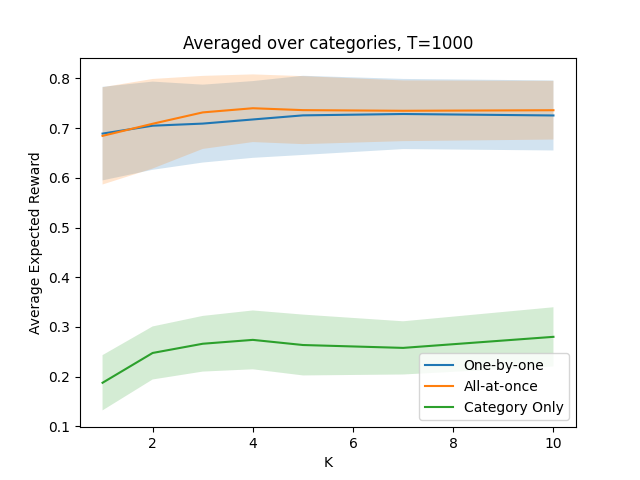}
        \label{fig:total_arxiv}
     \end{subfigure}
     \vspace{-5mm}
    \caption{Average expected reward $\AveRew(\text{category},K)$  (averaged over rounds and tasks), against $K$, the number of candidates. Each line corresponds to a prompting strategy or the Category-Only baseline.
    The shaded regions represent a $95\%$ confidence interval. A single arXiv category (``General Relativity and Quantum Cosmology", left), averages over 6 categories (right).}
    \label{fig:explore_arxiv}
\end{figure*}

\textbf{Explore puzzle: MovieLens.}
We run a MovieLens task for each of the $943$ users, with $K \in \{10, 19\}$. We re-run each task 5 times to denoise. For a given $K$, we record $\AveRew(K)$: reward $\RawRew$, defined above, averaged over all users and runs.
For baselines, we compare against running UCB1 for $T$ rounds over (1) a random subset of $K$ movies and (2) a random movie from each of the $19$ movie genres in the dataset. (Different random draws for each user.)

Our results for $\mistral$ and $\qwen$ are in~\Cref{tab:movielens}
%
Both LLMs significantly outperform all baselines for both values of $K$.
We also tried running the experiment using $\llama$ and $\gemma$, but the list of all movies was too long to fit in the context window.

\begin{table}
\centering
\caption{Average Reward for $\qwen$, $\mistral$, and Baselines for $K \in \{10, 19\}$ in the MovieLens Task.\looseness-1}
\begin{tabular}{l r r}
\hline
 $\AveRew$ &  $K=10$ & $K=19$ \\
\hline
$\qwen$             & $671.88$ & $\mathbf{678.25}$ \\
$\mistral$          & $\mathbf{735.38}$ & $675.16$ \\
Random & $570.39$ & $599.73$ \\
Genre-based      & ---    & $617.27$ \\
\hline
\end{tabular}
\label{tab:movielens}
\end{table}

The LLM has implicit knowledge about each movie
which it may leverage when selecting the discretization.
To mitigate overfitting to “popular” movies, we prompted the LLM
to not take movie popularity into account. However,
this change did not significantly affect the rewards.

\begin{table}
\caption{Performance using all-at-once prompting on select arXiv tasks. Results are averaged over $10$ abstract-title pairs. The full table (\Cref{table-aao-arxiv}) may be found in the Appendix.\looseness-1}
\label{table-aao-arxiv-main}
\centering
\begin{tabular}{lcccc}
\toprule
 & $K=1$ & $K=2$ & $K=5$ & MMR ($K=5$)\\
\midrule
gr-qc & 0.63 & 0.64 & \textbf{0.68} & 0.40\\
hep-ex & 0.81 & 0.81 & \textbf{0.83} & 0.52\\
hep-lat & \textbf{0.72} & \textbf{0.72} & \textbf{0.72} & 0.43\\
hep-th & 0.65 & 0.71 & \textbf{0.73} & 0.40\\
math-ph & 0.64 & 0.73 & \textbf{0.74} & 0.39\\
nucl-ex & 0.73 & \textbf{0.79} & 0.75 & 0.41\\
\bottomrule
\end{tabular}
\end{table}

\xhdr{Explore puzzle: arXiv.}
%
Using the arXiv API~\citep{arxiv_api}, we collected $10$ abstract-title pairs from each of the $41$ arXiv categories.%
\footnote{The knowledge cutoffs for our models were in late 2023, and we only use papers uploaded after June 2024.}
Each abstract-title pair yields a task, with $K=1,2,5$ suggestions. Each task is re-run 5 times to denoise. Given $K$, we record  $\AveRew(\text{category},K)$: reward $\RawRew$, as defined above, averaged over all tasks within the same arXiv category and all runs.

We evaluate $\gptfouro$ as an exploration oracle for these tasks.
%
%
We visualize our findings in \Cref{table-aao-arxiv-main}
and \Cref{fig:explore_arxiv}.
\footnote{\Cref{fig:explore_arxiv} focuses on 6 categories: General relativity and quantum cosmology; computer vision and pattern recognition; statistics theory; biomolecules; signal processing; general economics.}
%

To assess the LLM's ability to specialize to a task, we consider a baseline where the candidate arms are generated by $\gptfouro$ given only the category, \emph{not the abstract} (\Cref{table-co-arxiv}).
We also consider a stronger semantic baseline: Maximal Marginal Relevance (MMR) baseline~\citep{carbonell1998use}, which uses a retrieval and diversification pipeline over the arXiv paper pool we ran experiments on.
For each evaluation instance, the true title is removed from the pool so it cannot appear as an arm; remaining papers are scored by cosine similarity between the abstract embedding and each title embedding, and MMR picks $K$ titles by trading off between relevance and similarity to the other selected titles to balance topical match with diversity.
We then run the same UCB1 algorithm, with these titles as arms.

%

\xhdr{Explore puzzle: Q/A task.}
We used $\gptfour$ to generate a dataset of $10$ open-ended questions with many reasonable answers, along with an intentionally contrarian answer for each question to serve as the ground truth. (\Eg ``What does it mean to live a fulfilling life?'' ``Fulfillment comes from embracing discomfort.'')
Each question-answer pair yields a task, with $K \in \{1, 2, 3, 4, 5, 7, 10\}$ candidate answers. For a given $K$, we run each task 10 times and record
    $\AveRew(\text{task},K)$: reward $\RawRew$ averaged over all runs.

We evaluate $\gptfouro$, $\qwen$, $\gemma$, and $\mistral$ as exploration oracles. We also consider a non-LLM baseline (\term{Random}), which picks
$K$ arms independently u.a.r. in the embedding space (leveraging the AI-based embedding but suffering from its high dimension).%

The LLMs perform similarly, with $\AveRew\approx$0.5-0.6.
and typically work best for $K$=3,4.
see
\Cref{sec:QA_app}. In contrast, \term{Random}  catastrophically fails, with $\AveRew<0.1$. Moreover, we observe that LLM-generated answers pass the ``eye test'' (they are reasonable, yet spiritually and semantically different), and yield substantially different expected rewards.  We conclude that the LLM does succeed as an exploration oracle for these tasks.

%% file: main/conc.tex
\section{Conclusions}

Our work adds to (and provides partial explanations for) the growing literature  on in-context learning for decision-making tasks.
%
We find that LLMs 
are useful as \emph{exploration oracles} that propose high-quality candidate actions 
in large, semantically meaningful action spaces.
%
However, current LLMs are not that good at 
\emph{exploitation},
particularly in larger or more complex tasks, although reasoning models (specifically, the \textsc{Gpt-5} family of models) show the most promise. 
While we experiment with several mitigations including tool use and in-context summaries, they consistently underperform relative to a simple linear regression baseline, even in inherently non-linear tasks.\looseness-1
%
%
%

We provide a systematic evaluation of LLMs' capabilities as exploration oracles, in isolation from other components of a decision-making agent. We find robust success across several different task types, LLMs, and prompting strategies. In contrast, non-LLM baselines are often quite weak, absent known problem structure to be leveraged by an algorithm.

LLMs' continued difficulties at exploitation tasks should be contrasted with the rapid progress in LLMs and the extreme optimism associated with this progress. Ever since the $\gptthree$ release in 2022 (and especially since $\gptfour$ frenzy of early 2023), it was commonly believed that ``generalist'' frontier LLMs will soon master many/most relatively simple tasks: if not yet, then more likely in several months than in several years. However, our results show that these difficulties persist longer than anticipated.

\xhdr{Future work.}
%
LLM-based exploration oracles may potentially help with ``smart discretization" for rich text-based action spaces. Non-LLM bandit algorithms tend to fail in such scenarios, as discussed in \Cref{sec:explore}. 
The hope is to ``zoom in" entirely in the space of ``potentially relevant" actions, rather than in the space of \emph{all} actions.

Second, while we experiment with tool use for exploitation tasks, we find that it does not improve performance across the board. 
One could investigate various approaches for tool use in these tasks, and 
whether/when they help. 

%% file: app/QA_app.tex
Here we provide the key plots for the Q/A explore task, as defined in \Cref{sec:explore}. As stated there, all LLMs we considered perform similarly, with $\AveRew\approx$0.5-0.6.
and typically work best for $K$=3,4, whereas the \term{Random} baseline catastrophically fails, with $\AveRew<0.1$. Recall that 
    $\AveRew(\text{task},K)$
is the average reward $\RawRew$ for a given task with a given $K$ (as defined in \Cref{sec:explore}), averaged over all runs of this task.

\begin{figure*}[h]
     \centering
     \begin{subfigure}[b]{0.49\textwidth}
        \centering
        \includegraphics[width=\linewidth]{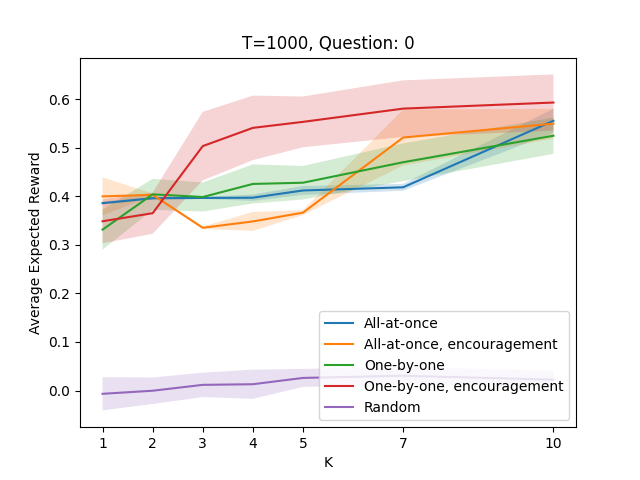}
        \label{fig:explore_q0}
     \end{subfigure}
     \hfill
     \begin{subfigure}[b]{0.49\textwidth}
        \centering
        \includegraphics[width=\linewidth]{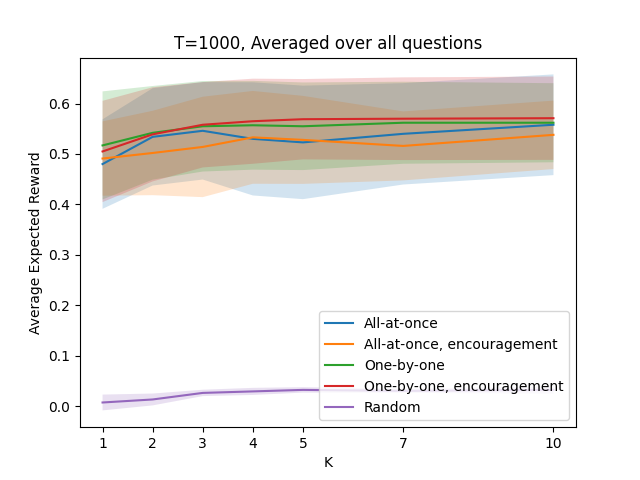}
        \label{fig:explore_all}
     \end{subfigure}
    \vspace{-5mm}
    \caption{Average expected reward $\AveRew(\text{task},K)$  (averaged over rounds and over runs), against $K$, \#candidates. Each line corresponds to a prompting strategy or the \term{Random} baseline.
    The shaded regions represent a $95\%$ confidence interval.}
    \label{fig:ucb}
\end{figure*}

We visualize our findings in \Cref{fig:ucb}. We plot $\AveRew(\text{task},K)$ against $K$, for one  particular task (left) and averaged across all tasks (right). Similar plots for the $9$ other tasks can be found in \Cref{app:QA-additional}.

\begin{figure*}[h]
     \centering
     \begin{subfigure}[b]{0.49\textwidth}
        \centering
        \includegraphics[width=\linewidth]{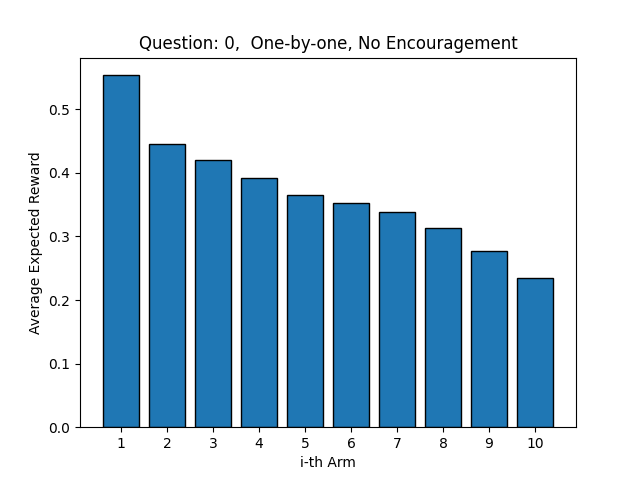}
        \label{fig:hist_q0}
     \end{subfigure}
     \hfill
     \begin{subfigure}[b]{0.49\textwidth}
        \centering
        \includegraphics[width=\linewidth]{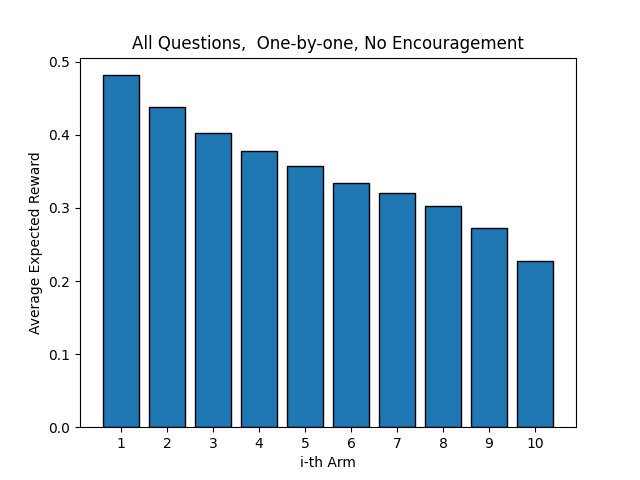}
        \label{fig:hist_all}
     \end{subfigure}
     \vspace{-5mm}
    \caption{Spread of expected rewards within the candidate set. Arm histogram for one particular question (``What is the meaning of freedom?'', left) and averaged over all questions (right). We consider $K=10$ suggestions generated by our one-by-one prompt without encouragement. For a given ``run", the $K$ suggestions are ranked by the expected reward, and then the $i$-th expected reward is averaged over all runs, for each $i\in [K]$.}
    \label{fig:hist}
\end{figure*}

The LLM-generated answers pass the ``eye test'': they are reasonable, yet spiritually and semantically different. 
E.g., given the question ``What is the role of technology in society'', the first $K=5$ suggestions generated by our 1-by-1 prompt are: (1) Facilitates communication, innovation, and efficiency, (2) Transforms daily life and shapes culture, (3) Drives connectivity and enhances productivity, (4) Facilitates control and surveillance, (5) Disrupts traditional relationships and norms. Finally, we verify that the candidate suggestions are substantially \emph{different} from one another in terms of their rewards, see \Cref{fig:hist}. 


%% file: app/exploit_app.tex
\subsection{Prompts}
\label[appendix]{app:prompts}

In this section we give example prompts for each of our experimental setups.

``Buttons'' prompt for the MAB puzzle:
\begin{quote}
    [SYSTEM] You are in a room with 5 buttons labeled blue, green, red, yellow, purple. Each button is associated with a Bernoulli distribution with a fixed but unknown mean; the means for the buttons could be different. For each button, when you press it, you will get a reward that is sampled from the button's associated distribution. Then you must pick the button with the highest empirical average, which must be exactly one of blue, green, red, yellow, or purple. You must provide your final answer immediately within the tags \textless Answer\textgreater COLOR \textless Answer\textgreater where COLOR is one of blue, green, red, yellow, or purple and with no text explanation.

    [USER] The past rewards for each button are:

    round $1$: blue button had reward $1$, green button had reward $1$, red button had reward $0$, yellow button had reward $1$, purple button had reward $0$

    round $2$: blue button had reward $0$, green button had reward $1$, red button had reward $1$, yellow button had reward $1$, purple button had reward $0$

    Which button do you choose? Remember, YOU MUST provide your final answer within the tags \textless Answer\textgreater COLOR \textless Answer\textgreater where COLOR is one of blue, green, red, yellow, or purple and with no text explanation.
\end{quote}

``Adverts'' prompt for the MAB puzzle:
\begin{quote}
    [SYSTEM] You are recommendation engine that chooses advertisements to display to users when they visit your webpage. There are 5 advertisements you can choose from, named A, B, C, D, E. When a user visits the webpage you can choose an advertisement to display and you will observe whether the user would have clicked each of the ads. You model this by assuming that each advertisement has a certain click rate and users click on advertisements with their corresponding rates. I will show you the past clicks for each advertisement. Then you must pick the advertisement with the highest empirical click rate, which must be exactly one of A, B, C, D, or E. You must provide your final answer immediately and with no text explanation. within the tags \textless Answer\textgreater ADVERTISEMENT \textless Answer\textgreater where ADVERTISEMENT is one of A, B, C, D, or E.

    [USER] The past clicks for each advertisement are:

    round $1$: advertisement A was clicked, advertisement B was clicked, advertisement C was not clicked, advertisement D was clicked, advertisement E was clicked

    round $1$: advertisement A was not clicked, advertisement B was clicked, advertisement C was clicked, advertisement D was clicked, advertisement E was not clicked

    Which advertisement do you choose? Remember, YOU MUST provide your final answer within the tags \textless Answer\textgreater ADVERTISEMENT \textless Answer\textgreater where ADVERTISEMENT is one of A, B, C, D, or E and with no text explanation.
\end{quote}

``Buttons'' prompt for the numerical CB puzzle:

\begin{quote}
    [SYSTEM] You are in a room with a television and 2 buttons labeled blue, green. Each button is associated with a Bernoulli distribution with an unknown mean; the means for the buttons could be different from each other and may depend on the list of numbers shown on the screen (i.e. the context). For each button, when you press it, you will get a reward that is sampled from the button's associated distribution, conditioned on the numbers shown on the television screen. I will show you the past numbers shown on the screen and the corresponding rewards for each button. A new list of numbers will then appear on the screen and you must pick the next button in order to maximize your reward in this round only, which must be exactly one of blue or green. You must provide your final answer immediately within the tags <Answer> COLOR </Answer> where COLOR is one of blue or green and with no text explanation.

    [USER] The past contexts and rewards for each button are:

    In round $1$, the context was $[0.3, 0.7]$. The blue button had reward $1$, the green button had reward $1$

    In round $2$, the context was $[0.4, 0.6]$. The blue button had reward $0$, the green button had reward $1$

    Which button do you choose? Remember, YOU MUST provide your final answer within the tags \textless Answer\textgreater COLOR \textless Answer\textgreater where COLOR is one of blue or green and with no text explanation.
\end{quote}

Prompt for the text-based CB puzzle:

\begin{quote}
    [SYSTEM] You are in a room with a table and a button. There may also be other objects in the room, which I will tell you about. You must then take one of the following actions: "pet animal", "leave room", "use tool", "eat food", "press button", after which you will receive some reward. The reward you receive is a random function of both the action you take and the information you receive about the objects in the room and time of day. Your goal is to maximize the expected reward you receive. I will show you the past history of play over 2 rounds. For each round, I will show you the state of the room and the corresponding rewards for each action. I will then tell you the current state of the room, and you must pick the next action in order to maximize your reward in this round only, which must be exactly one of "pet animal", "leave room", "use tool", "eat food", or "press button". Look for patterns in the data and try to estimate the reward of each action, given the information at your disposal. You must provide your final answer immediately within the tags \textless Answer\textgreater ACTION \textless Answer\textgreater where ACTION is one of "pet animal", "leave room", "use tool", "eat food", or "press button" and with no text explanation.

    [USER] The past observations and outcomes for each action are:

    Round $1$ had context time of day: morning, animal: bear, table item: chest, tool: key, food: apple, button color: red. "pet animal" had reward $0$, "leave room" had reward $1$, "use tool" had reward $1$, "eat food" had reward $0$, press button had reward $0$

    Round $2$ had context time of day: afternoon, animal: cat, table item: card, tool: hammer, food: cake, button color: orange. "pet animal" had reward $1$, "leave room" had reward $0$, "use tool" had reward $0$, "eat food" had reward $1$, press button had reward $0$

    The current state of the room is time of day: evening, animal: bear, table item: envelope, tool: key, food: nut, button color: red.

    Which action do you choose? Remember, you must provide your final answer immediately within the tags \textless Answer\textgreater ACTION \textless Answer\textgreater where ACTION is one of "pet animal", "leave room", "use tool", "eat food", or "press button" and with no text explanation.
\end{quote}

\clearpage
\subsection{Additional Results on MAB exploit puzzle}

\begin{figure}[h]
    \centering
    \includegraphics[width=0.7\linewidth]{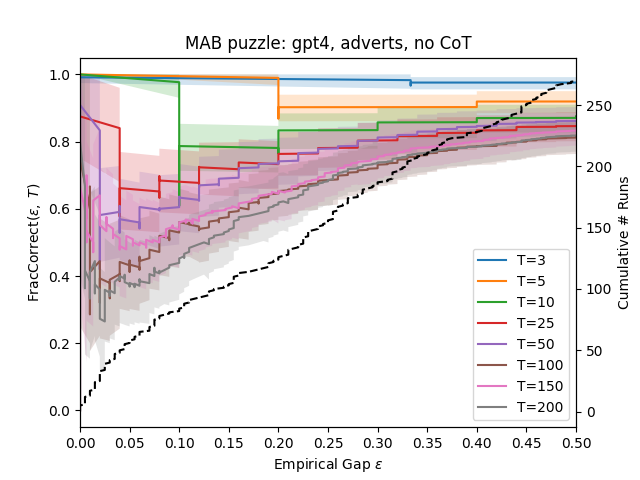}
    \caption{Cumulative fraction correct for $\gptfour$ in the MAB adverts puzzle.}
    \label{fig:mab_adverts}
\end{figure}
\begin{figure}[h]
    \centering
    \includegraphics[width=0.7\linewidth]{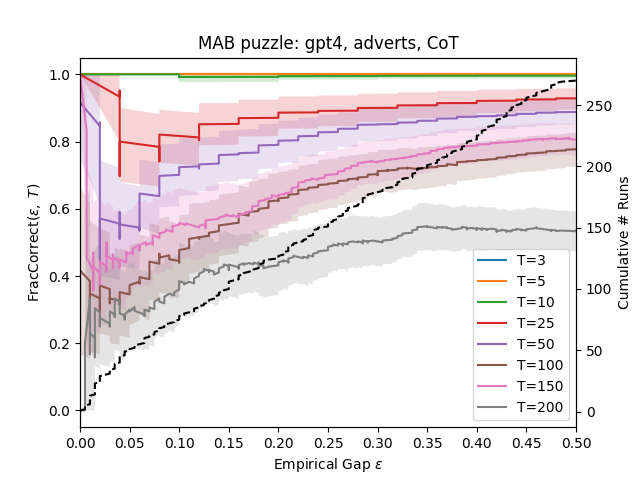}
      \caption{Cumulative fraction correct for $\gptfour$ with chain-of-thought reasoning in the MAB adverts puzzle.}
    \label{fig:mab_adverts_cot}
\end{figure}
\begin{figure}[h]
    \centering
    \includegraphics[width=0.7\linewidth]{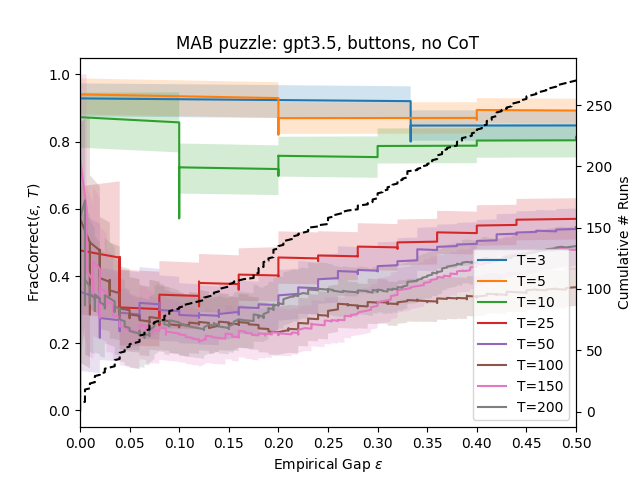}
    \caption{Cumulative fraction correct for $\gptthree$ in the MAB buttons puzzle.}
    \label{fig:mab_three_not_cot}
\end{figure}

\clearpage

\subsection{Additional details for text-based CB puzzles}

Each context contains a time of day (belonging to \{morning, afternoon, evening, night\}), an animal (\{bear, dog, cat, None\}), a tool (\{key, letter opener, hammer, None\}), a food item (\{cake, apple, nut, None\}), and a button with a particular color (\{red, orange, yellow, green\}).
The actions are ``pet animal'', ``leave room'', ``use tool'', ``eat food'', and ``press button''.

We considered two reward functions. One was ``easy'', with expected rewards (ExpRev) as follows:
\begin{itemize}
    \item ExpRev for petting the animal is 0.01 if the animal is a bear, 0.7 if the animal is a dog, and 0.4 if the animal is a cat. Otherwise, ExpRev if 0.5.
    \item ExpRev for leaving the room is always 0.5.
    \item ExpRev for using the tool is 0.75 if it is a key, 0.6 if it is a letter opener, 0.45 if it is a hammer, and 0.2 otherwise.
    \item ExpRev for eating food is 0.8 if it is cake, 0.6 if it is an apple, 0.2 if it is a nut, and 0.3 otherwise.
    \item ExpRev for pressing the button is 0.89 if it is green, 0.62 if it is yellow, 0.39 if it is orange, and 0.27 if it is red.
\end{itemize}

\begin{figure}[h]
    \centering
    \includegraphics[width=0.7\linewidth]{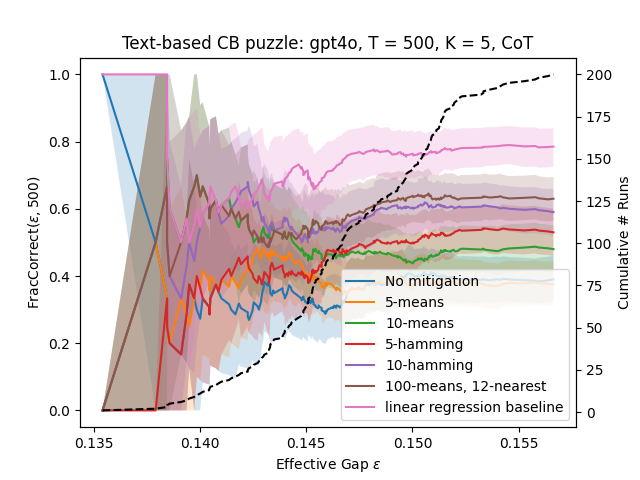}
    \caption{Performance of $\gptfouro$ with mitigations on the words CB puzzle with ``easy'' rewards.}
    \label{fig:words_easy}
\end{figure}

Our results under this reward function are summarized in~\Cref{fig:words_easy}. We used hamming distance to implement our mitigations. Note that in higher-dimensional settings, distance in an embedding space may be used.

The reward function we use in the main body is more complicated, and is detailed below:
\begin{itemize}
    \item ExpRev for petting the animal is 0.01 if it is a bear, 0.7 if it is a dog, 0.3 if it is a cat and the time of day is morning or afternoon, 0.7 if it is a cat and the time of day is evening or night, and otherwise 0.5.
    \item ExpRev for leaving the room is always 0.5
    \item If the animal is a bear, ExpRev for using the tool is 0.1. Otherwise, if the tool is a key and the table item is a chest, ExpRev is 0.9. Otherwise, it is 0.4
    \item If the animal is a bear, ExpRev for eating food is always 0.5. Otherwise, ExpRev is 0.8 for cake, 0.6 for an apple, 0.2 for a nut, and 0.5 otherwise.
    \item If the animal is a bear, ExpRev for pressing the button is 0.1. Otherwise if the button is green and the time of day is morning, or the button is yellow and the time of day is afternoon, or the button is orange and the time of day is evening, or the button is red and the time of day is night, then ExpRev is 0.9. In all other cases, ExpRev is 0.25.
\end{itemize}

\clearpage
\subsection{Robustness Check with Other Models}\label{sec:exploit-robustness}
In this appendix, we include results for Qwen2.5-7B-Instruct ($\qwen$), Gemma 3 12B ($\gemma$), and Mistral-7B-Instruct-v0.3 ($\mistral$) on our exploit puzzles.
Due to the time and costs associated with running our exploit puzzles, we prioritized (1) getting at least one additional set of results for each of our main puzzles (using $\qwen$) and (2) getting results for all of our models in our word-based puzzles.
Our findings are the same across all models.

\begin{figure}[h]
    \centering
    \includegraphics[width=0.7\linewidth]{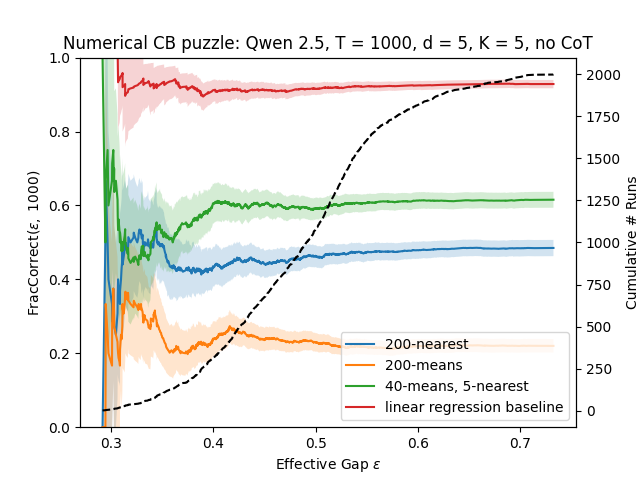}
    \caption{Results for $\qwen$ in the setting of~\Cref{fig:cb-main2}. Only mitigations are shown, as the unmitigated history is too long to fit in the context window.}
    \label{fig:1000-qwen}
\end{figure}

\begin{figure}[h]
    \centering
    \includegraphics[width=0.7\linewidth]{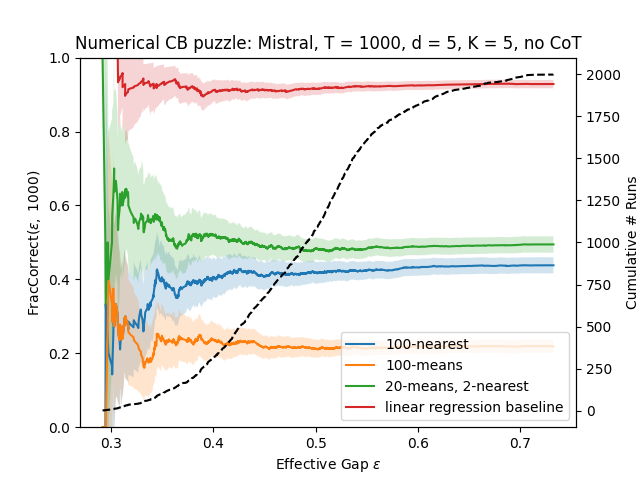}
    \caption{Results for $\mistral$ in the setting of~\Cref{fig:cb-main2}. Only mitigations are shown, as the unmitigated history is too long to fit in the context window.}
    \label{fig:1000-mistral}
\end{figure}

\begin{figure}[h]
    \centering
    \includegraphics[width=0.7\linewidth]{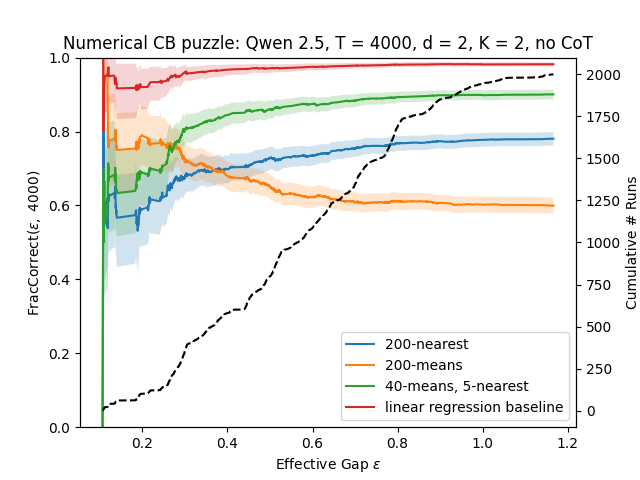}
    \caption{Results for $\qwen$ in the setting of~\Cref{fig:cb-main1}. Only mitigations are shown, as the unmitigated history is too long to fit in the context window.}
    \label{fig:4000-qwen}
\end{figure}

\begin{figure}[h]
    \centering
    \includegraphics[width=0.7\linewidth]{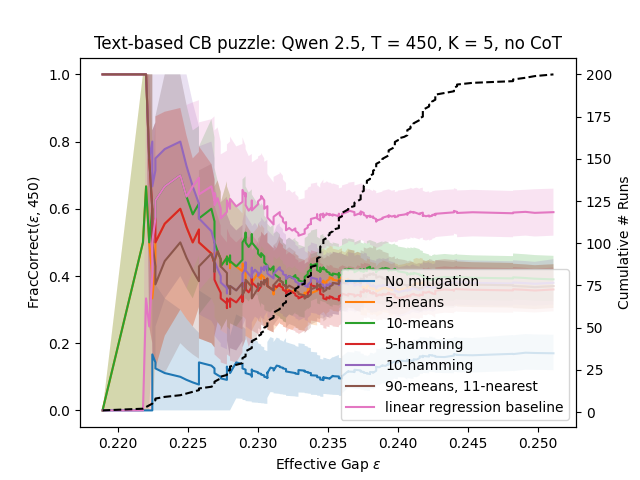}
    \caption{Results for $\qwen$ in the setting of~\Cref{fig:words} (right). We use a slightly smaller history length ($T=450$ instead of $T=500$) so that the full history can fit in the context window.}
    \label{fig:words-qwen-hard}
\end{figure}

\begin{figure}[h]
    \centering
    \includegraphics[width=0.7\linewidth]{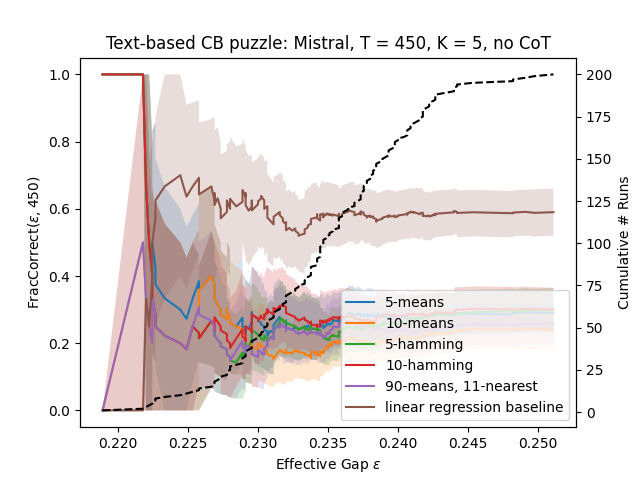}
    \caption{Results for $\mistral$ in the setting of~\Cref{fig:words-qwen-hard}. Only mitigations are shown, as the full history could not fit in the context window.}
    \label{fig:words-mistral-hard}
\end{figure}

\begin{figure}[h]
    \centering
    \includegraphics[width=0.7\linewidth]{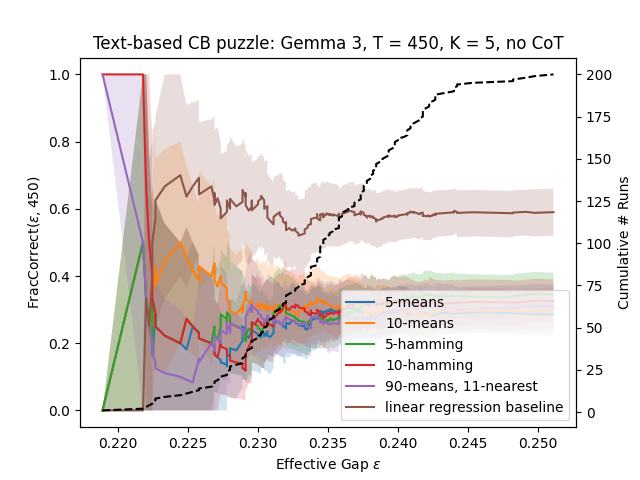}
    \caption{Results for $\gemma$ in the setting of~\Cref{fig:words-qwen-hard}. Only mitigations are shown, as the full history could not fit in the context window.}
    \label{fig:words-gemma-hard}
\end{figure}

\begin{figure}[h]
    \centering
    \includegraphics[width=0.7\linewidth]{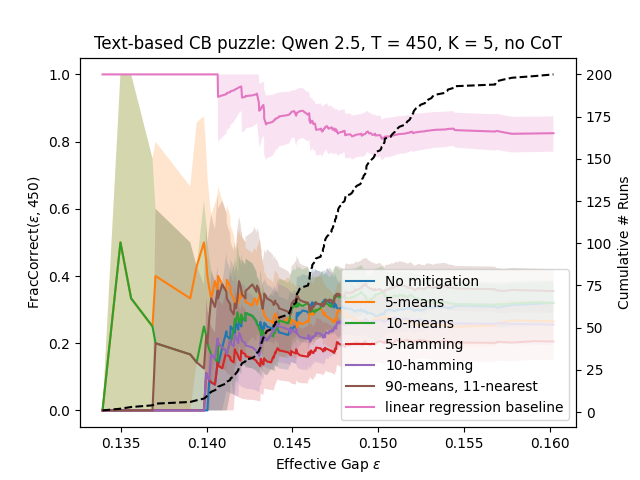}
    \caption{Results for $\qwen$ in the setting of~\Cref{fig:words_easy}. We use a slightly smaller history length ($T=450$ instead of $T=500$) so that the full history can fit in the context window.}
    \label{fig:words-qwen-easy}
\end{figure}

\begin{figure}[h]
    \centering
    \includegraphics[width=0.7\linewidth]{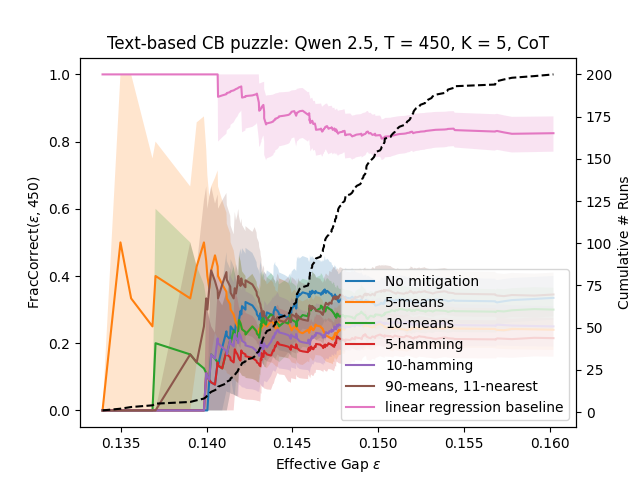}
    \caption{Results for $\qwen$ in the setting of~\Cref{fig:words_easy}, using chain-of-thought prompting. We use a slightly smaller history length ($T=450$ instead of $T=500$) so that the full history can fit in the context window.}
    \label{fig:words-qwen-easy-cot}
\end{figure}

\begin{figure}[h]
    \centering
    \includegraphics[width=0.7\linewidth]{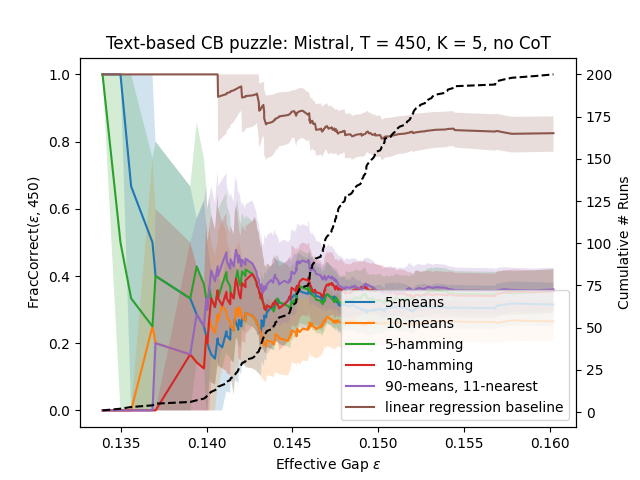}
    \caption{Results for $\mistral$ in the setting of~\Cref{fig:words-qwen-easy}. Only mitigations are shown, as the full history could not fit in the context window.}
    \label{fig:words-mistral-easy}
\end{figure}

\begin{figure}[h]
    \centering
    \includegraphics[width=0.7\linewidth]{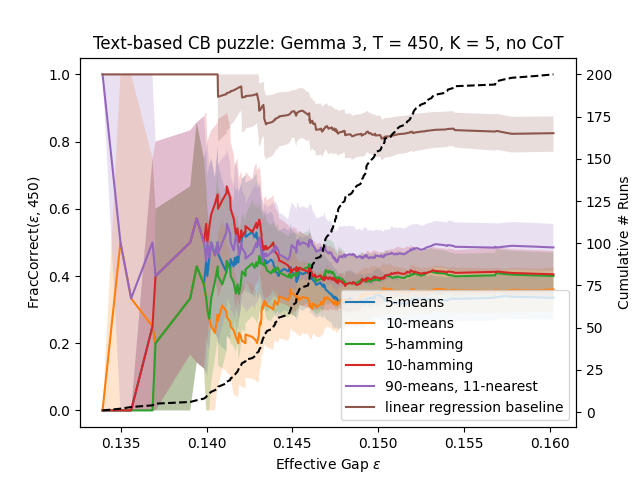}
    \caption{Results for $\gemma$ in the setting of~\Cref{fig:words-qwen-easy}. Only mitigations are shown, as the full history could not fit in the context window.}
    \label{fig:words-gemma-easy}
\end{figure}

\begin{figure}[h]
    \centering
    \includegraphics[width=0.7\linewidth]{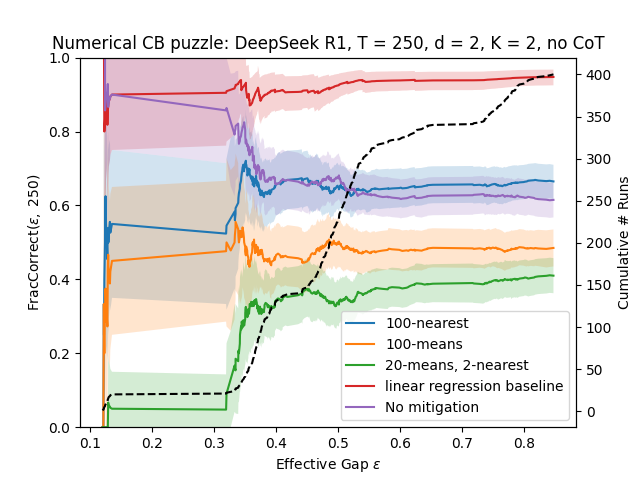}
    \caption{Performance of $\deepseek$ on our numerical CB puzzle.}
    \label{fig:placeholder}
\end{figure}

\clearpage

\subsection{Exploit Experiments using BanditBench}\label[appendix]{app:bb}

We construct an exploit puzzle using the CB setting in BanditBench~\citep{nie2024evolve} as a starting point. We modified the prompt to instruct the agent to exploit in each round.
Following their setup, in each round the agent is shown a feature vector consisting of both a textual description and numerical values corresponding to a user.
The goal is to give each user a personalized recommendation for a movie that they would be likely to enjoy.
Arms are movies, and rewards are constructed via a low rank approximation on the user-movie rating matrix from MovieLens.
For more details, see~\cite{nie2024evolve}.

Our findings carry over to this setting: we find that the $k$-nearest and $k$-nearest-means mitigations do help, but are still outperformed by the linear regression baseline. The details are as follows:


\begin{tabular}{l c c c c}
\hline
mitigations & none  & $10$-nearest & 10-means & $5$-nearest, $40$-means  \\ \hline
\gptfivetwo &$3.46$ & $3.83$ & $2.45$ & $3.57$ \\
\gptfivenano&$2.73$ & $3.86$ & $2.59$ & $3.65$ \\
Linear regression baseline & $4.06$
\end{tabular}

All of our results are for $T=200$ and are averaged over $500$ users. All rewards are deterministic and bounded in $[0, 5]$.

%% file: app/explore_app.tex
\subsection{Prompts}
\label[appendix]{app:explore-prompts}


Prompt for the MovieLens experiments:

\begin{quote}
    [SYSTEM] You are a movie expert helping a user choose a movie.

    [USER] Here is a list of movies with their numeric IDs: \ldots

    From this list, choose \{K\} movies. You don't know what taste in movies the user has, so select a diverse set of movies from different genres such that they will most likely enjoy at least one of the movies you select. Respond ONLY with the \{K\} numeric IDs, one per line, with no extra text. Do NOT consider a movie's popularity when deciding whether to select it.
\end{quote}

For the arXiv and Q/A tasks, we used a 2x2 space of prompts, as discussed in \Cref{sec:explore}: $\cbr{\text{all-at-once, 1-by-1}} \times \cbr{\text{with, without}}$ encouragement. We provide two illustrative examples below.

``All-at-once'' prompt with encouragement for the Q/A task:

\begin{quote}
    [SYSTEM] I will give you an open-ended question. Come up with 5 different candidate answers. Reply only with the 5 candidate answers, and put each candidate answer on a separate line. Each answer should only be a few words, skipping any introductory phrasing and going straight to the essence. Try to come up with answers that are very different in spirit from one another.

    [USER] Here is the question: ``What is the purpose of art?''
\end{quote}

``One-by-one'' prompt without encouragement for the arXiv task:

\begin{quote}
    [SYSTEM] I will give you an abstract and some candidate titles for a paper. Come up with a new candidate title that is relevant to the abstract, but different from the other candidate titles. Reply only with the candidate title.

    [USER] Here is the abstract: \{abstract goes here\}

    Here are the other candidate titles: \{previous suggestions go here\}
\end{quote}

\subsection{Dataset: Q/A task}
Our open-ended question dataset consists of the following $10$ questions and the corresponding ``ground-truth'' answers.

Questions:
\begin{enumerate}
    \setcounter{enumi}{-1}
    \item What is the meaning of freedom?
    \item How should we define success?
    \item What is the role of technology in society?
    \item What is the nature of reality?
    \item What is the purpose of art?
    \item What does it mean to live a fulfilling life?
    \item How do cultural differences shape our understanding of morality?
    \item What is the relationship between happiness and wealth?
    \item How can we balance individuality and community in modern society?
    \item What is the role of education in personal and societal growth?
\end{enumerate}

Answers:
\begin{enumerate}
    \setcounter{enumi}{-1}
    \item Freedom is an illusion shaped by societal norms and external influences.
    \item Success should be defined as contributing to the greater good rather than personal achievement.
    \item Technology disrupts the natural balance of society and often creates more problems than it solves.
    \item Reality is subjective, varying entirely based on individual perception and experience.
    \item The purpose of art is to challenge conventions and disrupt established ideas.
    \item Fulfillment comes from embracing discomfort.
    \item Cultural differences create moral superiority.
    \item Wealth detracts from true happiness.
    \item Individuality thrives when shaped by community.
    \item Education's purpose is to challenge authority.
\end{enumerate}

\subsection{Dataset: arXiv task}

Here is the list of paper titles we used in our arXiv dataset, along with their corresponding categories:

\begin{small}

gr-qc
\begin{enumerate}
\item There is more to the de Sitter horizon than just the area
\item Mitigating cosmic variance in the Hellings-Downs curve: a Cosmic
  Microwave Background analogy
\item Calabi-Yau Feynman integrals in gravity: $\varepsilon$-factorized form
  for apparent singularities
\item QG from SymQRG: AdS$_3$/CFT$_2$ Correspondence as Topological
  Symmetry-Preserving Quantum RG Flow
\item Black hole solutions in theories of supergravity
\item Horndeski in motion
\item Wormholes from beyond
\item Regularizing the Pulsar Timing Array likelihood: A path towards Fourier
  Space
\item Solutions to the mode equation for a quantized massless scalar field
  outside a black hole that forms from the collapse of a null shell: Late-time
  behaviors and computation of the stress-energy tensor
\item Gravitational waves from regular black holes in extreme mass-ratio
  inspirals
\end{enumerate}

hep-ex
\begin{enumerate}
\item Observation of the $K^{+} \rightarrow \pi^{+} \nu \bar\nu$ decay and
  measurement of its branching ratio
\item Test of lepton flavour universality in $W$-boson decays into electrons
  and $\tau$-leptons using $pp$ collisions at $\sqrt{s}=13$ TeV with the ATLAS
  detector
\item Searching for neutrino self-interactions at future muon colliders
\item Quantum Decoherence at ESSnuSB Experiment
\item Test of lepton flavour universality with $B^+ \to
  K^+\pi^+\pi^-\ell^+\ell^-$ decays
\item Cross-section measurements for the production of a $W$-boson in
  association with high-transverse-momentum jets in $pp$ collisions at
  $\sqrt{s}$= 13 TeV with the ATLAS detector
\item Charmful two-body $\Omega_b$ decays in the light-front quark model
\item Observation of a spectral hardening in cosmic ray boron spectrum with
  the DAMPE space mission
\item New BaBar studies of high-order radiation and the new landscape of
  data-driven HVP predictions of the muon g-2
\item Toponium: the smallest bound state and simplest hadron in quantum
  mechanics
\end{enumerate}

hep-lat
\begin{enumerate}
\item Quantum sampling on a quantum annealer for large volumes in the strong
  coupling limit for gauge group U(3)
\item Phase diagram of Rydberg atoms in a two-leg rectangular ladder
\item Graph Attention Hamiltonian Neural Networks: A Lattice System Analysis
  Model Based on Structural Learning
\item What do we know about the confinement mechanism?
\item Designing weight regularizations based on Lefschetz thimbles to
  stabilize complex Langevin
\item Likelihood of a zero in the proton elastic electric form factor
\item Real-Time Simulation of Asymmetry Generation in Fermion-Bubble
  Collisions
\item Investigating SU(3) with Nf=8 fundamental fermions at strong
  renormalized coupling
\item The determination of potential scales in 2+1 flavor QCD
\item Towards the phase diagram of fermions coupled with $SO(3)$ quantum links
  in $(2+1)$-D
\end{enumerate}

hep-ph
\begin{enumerate}
\item Predictions for dimuon production in high-energy neutrino-proton
  collisions using the color dipole model
\item Extrapolating Jet Radiation with Autoregressive Transformers
\item Accurate Surrogate Amplitudes with Calibrated Uncertainties
\item Calabi-Yau Feynman integrals in gravity: $\varepsilon$-factorized form
  for apparent singularities
\item The causal structure of the quark propagator
\item Fuzzy Axions and Associated Relics
\item Non-Radial Oscillation Modes in Hybrid Stars with Hyperons and Delta
  Baryons: Full General Relativity Formalism vs. Cowling Approximation
\item Evidence for the Sombrero Galaxy as an Accelerator of the Highest-Energy
  Cosmic Rays
\item The cosmic history of Primordial Black Hole accretion and its
  uncertainties
\item Searching for neutrino self-interactions at future muon colliders
\end{enumerate}

hep-th
\begin{enumerate}
\item There is more to the de Sitter horizon than just the area
\item Calabi-Yau Feynman integrals in gravity: $\varepsilon$-factorized form
  for apparent singularities
\item QG from SymQRG: AdS$_3$/CFT$_2$ Correspondence as Topological
  Symmetry-Preserving Quantum RG Flow
\item Geometrically constrained localized configurations engendering
  non-topological profile
\item The causal structure of the quark propagator
\item Entanglement Hamiltonian and orthogonal polynomials
\item Black hole solutions in theories of supergravity
\item Fuzzy Axions and Associated Relics
\item Celestial Mellin Amplitudes
\item Evidence for the Sombrero Galaxy as an Accelerator of the Highest-Energy
  Cosmic Rays
\end{enumerate}

math-ph
\begin{enumerate}
\item QG from SymQRG: AdS$_3$/CFT$_2$ Correspondence as Topological
  Symmetry-Preserving Quantum RG Flow
\item Entanglement Hamiltonian and orthogonal polynomials
\item Fermi's golden rule in tunneling models with quantum waveguides
  perturbed by Kato class measures
\item Semiclassical measure of the propagation between two topological
  insulators
\item On the Protection Against Noise for Measurement-Based Quantum
  Computation
\item Calculating Spectra by Sequential High-Pass Filtering
\item Validity of the stochastic Landau approximation for super-pattern
  forming systems with a spatial 1:3 resonance
\item Multi-component Hamiltonian difference operators
\item Emptiness Instanton in Quantum Polytropic Gas
\item Unitary $n$-correlations with restricted support in random matrix theory
\end{enumerate}

nucl-ex
\begin{enumerate}
\item The evidence of $N=16$ shell closure and $\beta$-delayed neutron
  emission from $\wedge^{25}$F
\item Isotopic Transparency in Central Xe+Sn Collisions at 100 MeV/nucleon
\item Detecting the Coupling of Axion Dark Matter to Neutron Spins at
  Spallation Sources via Rabi Oscillation
\item Likelihood of a zero in the proton elastic electric form factor
\item Nuclear structure and direct reaction studies in particle-$\gamma$
  coincidence experiments at the FSU John D. Fox Superconducting Linear
  Accelerator Laboratory
\item Bottomonium-like states in proton collisions: Fragmentation and
  resummation
\item Towards a foundation model for heavy-ion collision experiments through
  point cloud diffusion
\item Study of the energy spectrum of alpha particles in an experiment on
  irradiation of a boron target with a proton beam at the Prometheus
  accelerator
\item Staking out the Proton Drip-Line of Thulium at the N=82 Shell Closure
\item Measurements of global and local spin polarization of $\wedge$ and
  $\bar{\wedge}$ in Au+Au collisions from the RHIC Beam Energy Scan
\end{enumerate}

nucl-th
\begin{enumerate}
\item The causal structure of the quark propagator
\item Non-Radial Oscillation Modes in Hybrid Stars with Hyperons and Delta
  Baryons: Full General Relativity Formalism vs. Cowling Approximation
\item Isotopic Transparency in Central Xe+Sn Collisions at 100 MeV/nucleon
\item Quantum-Corrected Holographic Wilson Loop Correlators and Confinement
\item Dynamics of Hot QCD Matter 2024 -- Bulk Properties
\item Spurious Isospin Breaking in the In-medium Similarity Renormalization
  Group
\item Likelihood of a zero in the proton elastic electric form factor
\item Born-Oppenheimer Renormalization group for High Energy Scattering: the
  Modified BFKL, or where did it all go?
\item Nuclear structure and direct reaction studies in particle-$\gamma$
  coincidence experiments at the FSU John D. Fox Superconducting Linear
  Accelerator Laboratory
\item Bottomonium-like states in proton collisions: Fragmentation and
  resummation
\end{enumerate}

quant-ph
\begin{enumerate}
\item Noisy initial-state qubit-channel metrology with additional undesirable
  noisy evolution
\item The State Preparation of Multivariate Normal Distributions using Tree
  Tensor Network
\item QG from SymQRG: AdS$_3$/CFT$_2$ Correspondence as Topological
  Symmetry-Preserving Quantum RG Flow
\item Entanglement Hamiltonian and orthogonal polynomials
\item Learning interactions between Rydberg atoms
\item The isoholonomic inequality and tight implementations of holonomic
  quantum gates
\item Fermi's golden rule in tunneling models with quantum waveguides
  perturbed by Kato class measures
\item Temporal evolution of a forced optomechanical system with linear and
  quadratic field -- mechanical oscillator couplings
\item Control of a Josephson Digital Phase Detector via an SFQ-based Flux Bias
  Driver
\item Commentary on the decomposition of universal multiport interferometers:
  how it works in practice
\end{enumerate}

cs.AI
\begin{enumerate}
\item MaxInfoRL: Boosting exploration in reinforcement learning through
  information gain maximization
\item SepLLM: Accelerate Large Language Models by Compressing One Segment into
  One Separator
\item Stabilizing Reinforcement Learning in Differentiable Multiphysics
  Simulation
\item Revelations: A Decidable Class of POMDPs with Omega-Regular Objectives
\item Artificial Intelligence in Traffic Systems
\item The Impact of AI Assistance on Radiology Reporting: A Pilot Study Using
  Simulated AI Draft Reports
\item Can LLM Prompting Serve as a Proxy for Static Analysis in Vulnerability
  Detection
\item FSFM: A Generalizable Face Security Foundation Model via Self-Supervised
  Facial Representation Learning
\item Learning to Navigate in Mazes with Novel Layouts using Abstract Top-down
  Maps
\item SpeechPrune: Context-aware Token Pruning for Speech Information
  Retrieval
\end{enumerate}

cs.CL
\begin{enumerate}
\item SepLLM: Accelerate Large Language Models by Compressing One Segment into
  One Separator
\item Making FETCH! Happen: Finding Emergent Dog Whistles Through Common
  Habitats
\item Semi-automated analysis of audio-recorded lessons: The case of teachers'
  engaging messages
\item Virtual Agent-Based Communication Skills Training to Facilitate Health
  Persuasion Among Peers
\item How Private are Language Models in Abstractive Summarization?
\item Can LLM Prompting Serve as a Proxy for Static Analysis in Vulnerability
  Detection
\item SpeechPrune: Context-aware Token Pruning for Speech Information
  Retrieval
\item The Open Source Advantage in Large Language Models (LLMs)
\item LLM-RG4: Flexible and Factual Radiology Report Generation across Diverse
  Input Contexts
\item ExecRepoBench: Multi-level Executable Code Completion Evaluation
\end{enumerate}

cs.CV
\begin{enumerate}
\item PanSplat: 4K Panorama Synthesis with Feed-Forward Gaussian Splatting
\item Causal Diffusion Transformers for Generative Modeling
\item CAP4D: Creating Animatable 4D Portrait Avatars with Morphable Multi-View
  Diffusion Models
\item Wonderland: Navigating 3D Scenes from a Single Image
\item Stabilizing Reinforcement Learning in Differentiable Multiphysics
  Simulation
\item Instruction-based Image Manipulation by Watching How Things Move
\item IDArb: Intrinsic Decomposition for Arbitrary Number of Input Views and
  Illuminations
\item UniLoc: Towards Universal Place Recognition Using Any Single Modality
\item CPath-Omni: A Unified Multimodal Foundation Model for Patch and Whole
  Slide Image Analysis in Computational Pathology
\item CG-Bench: Clue-grounded Question Answering Benchmark for Long Video
  Understanding
\end{enumerate}

cs.LG
\begin{enumerate}
\item MaxInfoRL: Boosting exploration in reinforcement learning through
  information gain maximization
\item SepLLM: Accelerate Large Language Models by Compressing One Segment into
  One Separator
\item No More Tuning: Prioritized Multi-Task Learning with Lagrangian
  Differential Multiplier Methods
\item Stabilizing Reinforcement Learning in Differentiable Multiphysics
  Simulation
\item Extrapolating Jet Radiation with Autoregressive Transformers
\item Bilevel Learning with Inexact Stochastic Gradients
\item LLMs for Cold-Start Cutting Plane Separator Configuration
\item LeARN: Learnable and Adaptive Representations for Nonlinear Dynamics in
  System Identification
\item Thermodynamics-informed graph neural networks for real-time simulation
  of digital human twins
\item Memory-Reduced Meta-Learning with Guaranteed Convergence
\end{enumerate}

cs.NE
\begin{enumerate}
\item Deep-learning-based identification of individual motion characteristics
  from upper-limb trajectories towards disorder stage evaluation
\item Speeding Up the NSGA-II With a Simple Tie-Breaking Rule
\item Optimal Gradient Checkpointing for Sparse and Recurrent Architectures
  using Off-Chip Memory
\item Runtime Analysis for Multi-Objective Evolutionary Algorithms in
  Unbounded Integer Spaces
\item Theoretical Analysis of Quality Diversity Algorithms for a Classical
  Path Planning Problem
\item Populating cellular metamaterials on the extrema of attainable
  elasticity through neuroevolution
\item Deployment Pipeline from Rockpool to Xylo for Edge Computing
\item Interlocking-free Selective Rationalization Through Genetic-based
  Learning
\item EVOS: Efficient Implicit Neural Training via EVOlutionary Selector
\item Brain-inspired Chaotic Graph Backpropagation for Large-scale
  Combinatorial Optimization
\end{enumerate}

cs.RO
\begin{enumerate}
\item MaxInfoRL: Boosting exploration in reinforcement learning through
  information gain maximization
\item Stabilizing Reinforcement Learning in Differentiable Multiphysics
  Simulation
\item LeARN: Learnable and Adaptive Representations for Nonlinear Dynamics in
  System Identification
\item Backstepping Control of Tendon-Driven Continuum Robots in Large
  Deflections Using the Cosserat Rod Model
\item Learning to Navigate in Mazes with Novel Layouts using Abstract Top-down
  Maps
\item Emma-X: An Embodied Multimodal Action Model with Grounded Chain of
  Thought and Look-ahead Spatial Reasoning
\item Lightweight Decentralized Neural Network-Based Strategies for
  Multi-Robot Patrolling
\item Learning Human-Aware Robot Policies for Adaptive Assistance
\item Hardware-in-the-loop Simulation Testbed for Geomagnetic Navigation
\item Sonar-based Deep Learning in Underwater Robotics: Overview, Robustness
  and Challenges
\end{enumerate}

cs.IT
\begin{enumerate}
\item Codes from $A_m$-invariant polynomials
\item BA-BFL: Barycentric Aggregation for Bayesian Federated Learning
\item Capacity of Hierarchical Secure Coded Gradient Aggregation with
  Straggling Communication Links
\item Wireless Environmental Information Theory: A New Paradigm towards 6G
  Online and Proactive Environment Intelligence Communication
\item Quantum search in a dictionary based on fingerprinting-hashing
\item Identification Over Binary Noisy Permutation Channels
\item Iterative Detection and Decoding for Clustered Cell-Free Massive MIMO
  Networks
\item Structured Sampling for Robust Euclidean Distance Geometry
\item Study of Iterative Detection and Decoding for Multiuser Systems and MMSE
  Refinements with Active or Passive RIS
\item Shannon information and integrated information: message and meaning
\end{enumerate}

cs.CR
\begin{enumerate}
\item Can LLM Prompting Serve as a Proxy for Static Analysis in Vulnerability
  Detection
\item Efficient Layered New Bit-Flipping QC-MDPC Decoder for BIKE Post-Quantum
  Cryptography
\item But Can You Use It? Design Recommendations for Differentially Private
  Interactive Systems
\item Efficiently Achieving Secure Model Training and Secure Aggregation to
  Ensure Bidirectional Privacy-Preservation in Federated Learning
\item On Large Language Models in Mission-Critical IT Governance: Are We Ready
  Yet?
\item Just a Simple Transformation is Enough for Data Protection in Vertical
  Federated Learning
\item SeSeMI: Secure Serverless Model Inference on Sensitive Data
\item DB-PAISA: Discovery-Based Privacy-Agile IoT Sensing+Actuation
\item OTA-Key: Over the Air Key Management for Flexible and Reliable IoT
  Device Provision
\item Android App Feature Extraction: A review of approaches for malware and
  app similarity detection
\end{enumerate}

cs.DS
\begin{enumerate}
\item Approximating the Top Eigenvector in Random Order Streams
\item Witty: An Efficient Solver for Computing Minimum-Size Decision Trees
\item Adaptive Manipulation for Coalitions in Knockout Tournaments
\item Counting Butterflies over Streaming Bipartite Graphs with Duplicate
  Edges
\item Quantum search in a dictionary based on fingerprinting-hashing
\item Regularized Dikin Walks for Sampling Truncated Logconcave Measures,
  Mixed Isoperimetry and Beyond Worst-Case Analysis
\item Proportionally Fair Matching via Randomized Rounding
\item Logarithmic Positional Partition Interval Encoding
\item New results for the detection of bicliques
\item Deterministic Even-Cycle Detection in Broadcast CONGEST
\end{enumerate}

cs.HC
\begin{enumerate}
\item Virtual Agent-Based Communication Skills Training to Facilitate Health
  Persuasion Among Peers
\item The Impact of AI Assistance on Radiology Reporting: A Pilot Study Using
  Simulated AI Draft Reports
\item Combining Large Language Models with Tutoring System Intelligence: A
  Case Study in Caregiver Homework Support
\item But Can You Use It? Design Recommendations for Differentially Private
  Interactive Systems
\item LLMs Can Simulate Standardized Patients via Agent Coevolution
\item LLM-DaaS: LLM-driven Drone-as-a-Service Operations from Text User
  Requests
\item Private Yet Social: How LLM Chatbots Support and Challenge Eating
  Disorder Recovery
\item Task-Based Role-Playing VR Game for Supporting Intellectual Disability
  Therapies
\item Privacy-Preserving Brain-Computer Interfaces: A Systematic Review
\item Accurate, Robust and Privacy-Preserving Brain-Computer Interface
  Decoding
\end{enumerate}

math.AG
\begin{enumerate}
\item Geometry of 3-dimensional del Pezzo fibrations in positive
  characteristic
\item The Mordell-Schinzel conjecture for cubic diophantine equations
\item The many faces of a logarithmic scheme
\item Lorentzian polynomials and the incidence geometry of tropical linear
  spaces
\item $p$-adic Local Langlands Correspondence
\item Real del Pezzo surfaces without points
\item Linearization problem for finite subgroups of the plane Cremona group
\item Groupes de monodromie finie des variétés abéliennes
\item Duality for Arithmetic $p$-adic Pro-étale Cohomology of Analytic
  Spaces
\item The external activity complex of a pair of matroids
\end{enumerate}

math.AT
\begin{enumerate}
\item Digital $n-$Manifolds With Or Without Boundaries
\item Spatiotemporal Persistence Landscapes
\item Simplifications of finite spaces equipped with sheaves
\item Rational homotopy theory of operad modules through colored operads
\item Algebraic Topology Without Open Sets: A Net Approach to Homotopy Theory
  in Limit Spaces
\item The geometry of simplicial distributions on suspension scenarios
\item On the Last Kervaire Invariant Problem
\item Machine Proofs for Adams Differentials and Extension Problems among CW
  Spectra
\item Finite asymptotic dimension and the coarse assembly map
\item Modeling $(\infty,1)$-categories with Segal spaces
\end{enumerate}

math.AP
\begin{enumerate}
\item Decay estimates for massive Dirac equation in a constant magnetic field
\item Semiclassical measure of the propagation between two topological
  insulators
\item Convex waves grazing convex obstacles to high order
\item A Note on Hyperbolic Relaxation of the Navier-Stokes-Cahn-Hilliard
  system for incompressible two-phase flow
\item Positive solutions to general semilinear overdetermined boundary
  problems
\item Capacitary measures in fractional order Sobolev spaces: Compactness and
  applications to minimization problems
\item Validity of the stochastic Landau approximation for super-pattern
  forming systems with a spatial 1:3 resonance
\item Spectral bounds for the operator pencil of an elliptic system in an
  angle
\item Infinite dimensional invariant tori for nonlinear Schrödinger
  equations
\item A Serrin-type over-determined problem for Hessian equations in the
  exterior domain
\end{enumerate}

math.CT
\begin{enumerate}
\item Open Condensed Subgroups and Mackey's Formula
\item The Relational Quotient Completion
\item Classification of localizing subcategories along t-structures
\item Categorification of modules and construction of schemes
\item Rational RG flow, extension, and Witt class
\item Intrinsically Correct Sorting in Cubical Agda
\item Single and multi-valued Hilbert-bundle renormings
\item Extended (tri)dendriform algebras, pre-Lie algebras and post-Lie
  algebras as companion structures of extended Rota-Baxter algebras
\item On The Telescopic Picard Group
\item Enhanced 2-categorical structures, two-dimensional limit sketches and
  the symmetry of internalisation
\end{enumerate}

math.GR
\begin{enumerate}
\item $F$-birestriction monoids in enriched signature
\item Linearization problem for finite subgroups of the plane Cremona group
\item Salter's question on the image of the Burau representation of $B_4$
\item Averaging groups
\item Enumerating Diagonalizable Matrices over $\mathbb{Z}_{p^k}$
\item The scale function for locally compact groups acting on non-positively
  curved spaces
\item A computational study of certain Weyl modules for type $G_2$ in
  characteristic 2
\item Left-Invariant Riemannian Distances on Higher-Rank Sol-Type Groups
\item Growth Rate Gap for Stable Subgroups
\item Computing Young's Natural Representations for Generalized Symmetric
  Groups
\end{enumerate}

math.NT
\begin{enumerate}
\item The Mordell-Schinzel conjecture for cubic diophantine equations
\item Simultaneous and multiplicative Diophantine approximation on
  missing-digit fractals
\item Codes from $A_m$-invariant polynomials
\item Generalised Fermat equation: a survey of solved cases
\item Groupes de monodromie finie des variétés abéliennes
\item Vanishing of Witten zeta function at negative integers
\item Popa's "Recurrent Sequences" and Reciprocity
\item Duality for Arithmetic $p$-adic Pro-étale Cohomology of Analytic
  Spaces
\item About Eisenstein's Theorem
\item On the packing dimension of weighted singular matrices on fractals
\end{enumerate}

math.OC
\begin{enumerate}
\item Bilevel Learning with Inexact Stochastic Gradients
\item Memory-Reduced Meta-Learning with Guaranteed Convergence
\item On Differential Stability of a Class of Convex Optimization Problems
\item Convergence of trust-region algorithms in compact metric spaces
\item Eckstein-Ferris-Pennanen-Robinson duality revisited: paramonotonicity,
  total Fenchel-Rockallar duality, and the Chambolle-Pock operator
\item Capacitary measures in fractional order Sobolev spaces: Compactness and
  applications to minimization problems
\item A monotone block coordinate descent method for solving absolute value
  equations
\item Bivariate rational approximations of the general temperature integral
\item Toward a Unified Theory of Gradient Descent under Generalized Smoothness
\item A particle system approach towards the global well-posedness of master
  equations for potential mean field games of control
\end{enumerate}

math.ST
\begin{enumerate}
\item Optimality of the Right-Invariant Prior
\item The entropic optimal (self-)transport problem: Limit distributions for
  decreasing regularization with application to score function estimation
\item Causal Invariance Learning via Efficient Optimization of a Nonconvex
  Objective
\item A partial likelihood approach to tree-based density modeling and its
  application in Bayesian inference
\item Dual Unscented Kalman Filter Architecture for Sensor Fusion in Water
  Networks Leak Localization
\item Learning Massive-scale Partial Correlation Networks in Clinical
  Multi-omics Studies with HP-ACCORD
\item Well-Posedness and Stability of the Stochastic OGTT Model
\item Posterior asymptotics of high-dimensional spiked covariance model with
  inverse-Wishart prior
\item Model checking for high dimensional generalized linear models based on
  random projections
\item The Stein-log-Sobolev inequality and the exponential rate of convergence
  for the continuous Stein variational gradient descent method
\end{enumerate}

q-bio.BM
\begin{enumerate}
\item Category-Specific Topological Learning of Metal-Organic Frameworks
\item Applications of Knot Theory for the Improvement of the AlphaFold Protein
  Database
\item EquiFlow: Equivariant Conditional Flow Matching with Optimal Transport
  for 3D Molecular Conformation Prediction
\item FlowDock: Geometric Flow Matching for Generative Protein-Ligand Docking
  and Affinity Prediction
\item NeuralPLexer3: Physio-Realistic Biomolecular Complex Structure
  Prediction with Flow Models
\item COMET: Benchmark for Comprehensive Biological Multi-omics Evaluation
  Tasks and Language Models
\item Quadratic unconstrained binary optimization and constraint programming
  approaches for lattice-based cyclic peptide docking
\item High-dimensional Statistics Applications to Batch Effects in
  Metabolomics
\item Precise Antigen-Antibody Structure Predictions Enhance Antibody
  Development with HelixFold-Multimer
\item Sampling-based Continuous Optimization with Coupled Variables for RNA
  Design
\end{enumerate}

q-bio.GN
\begin{enumerate}
\item BarcodeMamba: State Space Models for Biodiversity Analysis
\item VEPerform: a web resource for evaluating the performance of variant
  effect predictors
\item A robust, scalable K-statistic for quantifying immune cell clustering in
  spatial proteomics data
\item Can linguists better understand DNA?
\item A Misclassification Network-Based Method for Comparative Genomic
  Analysis
\item DNA Fragments in Crude Oil Reveals Earth's Hidden History
\item Ancient DNA from 120-Million-Year-Old Lycoptera Fossils Reveals
  Evolutionary Insights
\item Emerging Challenges in Molecular Paleontology: Misapplication of
  Environmental DNA Fragments and Misconception of Deamination as a Key
  Criterion for In Situ DNA Identification
\item ProtGO: A Transformer based Fusion Model for accurately predicting Gene
  Ontology (GO) Terms from full scale Protein Sequences
\item DART-Eval: A Comprehensive DNA Language Model Evaluation Benchmark on
  Regulatory DNA
\end{enumerate}

q-bio.QM
\begin{enumerate}
\item Deep-learning-based identification of individual motion characteristics
  from upper-limb trajectories towards disorder stage evaluation
\item Decoding Drug Discovery: Exploring A-to-Z In silico Methods for
  Beginners
\item BarcodeMamba: State Space Models for Biodiversity Analysis
\item FlowDock: Geometric Flow Matching for Generative Protein-Ligand Docking
  and Affinity Prediction
\item Reliable and superior elliptic Fourier descriptor normalization and its
  application software ElliShape with efficient image processing
\item MEATRD: Multimodal Anomalous Tissue Region Detection Enhanced with
  Spatial Transcriptomics
\item Cardiovascular Disease Detection By Leveraging Semi-Supervised Learning
\item Predictive Pattern Recognition Techniques Towards Spatiotemporal
  Representation of Plant Growth in Simulated and Controlled Environments: A
  Comprehensive Review
\item RAID-Database: human Responses to Affine Image Distortions
\item MiCull2 -- simulating mastitis transmission through milking order
\end{enumerate}

q-bio.PE
\begin{enumerate}
\item Asymmetric Interactions Shape Survival During Population Range
  Expansions
\item Quasispecies dynamics with time lags and periodic fluctuations in
  replication
\item Explicit modeling of density dependence in spatial capture-recapture
  models
\item Stochastic models in phylogenetic comparative methods: analytical
  properties and parameter estimation
\item Multivariate Aspects of Phylogenetic Comparative Methods
\item The expensive son hypothesis
\item Self-similarity in pandemic spread and fractal containment policies
\item Estimating excess mortality during the Covid-19 pandemic in Aotearoa New
  Zealand
\item An assessment of Alberta's strategy for controlling mountain pine beetle
  outbreaks
\item Mountain pine beetle struggles with jack pine: A mechanistic explanation
  for slowed range expansion in Alberta
\end{enumerate}

q-fin.CP
\begin{enumerate}
\item S\&P 500 Trend Prediction
\item Simulation of square-root processes made simple: applications to the
  Heston model
\item From Votes to Volatility Predicting the Stock Market on Election Day
\item SusGen-GPT: A Data-Centric LLM for Financial NLP and Sustainability
  Report Generation
\item FinGPT: Enhancing Sentiment-Based Stock Movement Prediction with
  Dissemination-Aware and Context-Enriched LLMs
\item Reciprocity in Interbank Markets
\item Integrative Analysis of Financial Market Sentiment Using CNN and GRU for
  Risk Prediction and Alert Systems
\item Financial Fine-tuning a Large Time Series Model
\item Geometric Deep Learning for Realized Covariance Matrix Forecasting
\item Isogeometric Analysis for the Pricing of Financial Derivatives with
  Nonlinear Models: Convertible Bonds and Options
\end{enumerate}

q-fin.PM
\begin{enumerate}
\item Cost-aware Portfolios in a Large Universe of Assets
\item PolyModel for Hedge Funds' Portfolio Construction Using Machine Learning
\item Geometric Deep Learning for Realized Covariance Matrix Forecasting
\item LLMs for Time Series: an Application for Single Stocks and Statistical
  Arbitrage
\item A Joint Energy and Differentially-Private Smart Meter Data Market
\item Smart leverage? Rethinking the role of Leveraged Exchange Traded Funds
  in constructing portfolios to beat a benchmark
\item Correlation without Factors in Retail Cryptocurrency Markets
\item Turnover of investment portfolio via covariance matrix of returns
\item MILLION: A General Multi-Objective Framework with Controllable Risk for
  Portfolio Management
\item Dynamic ETF Portfolio Optimization Using enhanced Transformer-Based
  Models for Covariance and Semi-Covariance Prediction(Work in Progress)
\end{enumerate}

q-fin.TR
\begin{enumerate}
\item Auto-Regressive Control of Execution Costs
\item FinGPT: Enhancing Sentiment-Based Stock Movement Prediction with
  Dissemination-Aware and Context-Enriched LLMs
\item Efficient and Verified Continuous Double Auctions
\item A Joint Energy and Differentially-Private Smart Meter Data Market
\item A theory of passive market impact
\item Uncertain Regulations, Definite Impacts: The Impact of the US Securities
  and Exchange Commission's Regulatory Interventions on Crypto Assets
\item Ergodic optimal liquidations in DeFi
\item MarketGPT: Developing a Pre-trained transformer (GPT) for Modeling
  Financial Time Series
\item Calculating Profits and Losses for Algorithmic Trading Strategies: A
  Short Guide
\item Market Making without Regret
\end{enumerate}

stat.AP
\begin{enumerate}
\item But Can You Use It? Design Recommendations for Differentially Private
  Interactive Systems
\item Efficient Bayesian inversion for simultaneous estimation of geometry and
  spatial field using the Karhunen-Loève expansion
\item Chopin: An Open Source R-language Tool to Support Spatial Analysis on
  Parallelizable Infrastructure
\item Spatial Cross-Recurrence Quantification Analysis for Multi-Platform
  Contact Tracing and Epidemiology Research
\item P3LS: Point Process Partial Least Squares
\item Missing data imputation for noisy time-series data and applications in
  healthcare
\item Balancing Accuracy and Costs in Cross-Temporal Hierarchies:
  Investigating Decision-Based and Validation-Based Reconciliation
\item Statistical Problems in the Diagnosis of Shaken Baby Syndrome/Abusive
  Head Trauma: Limitations to Algorithms and the Need for Reliable Data
\item CESAR: A Convolutional Echo State AutoencodeR for High-Resolution Wind
  Forecasting
\item Cardiovascular Disease Detection By Leveraging Semi-Supervised Learning
\end{enumerate}

stat.ML
\begin{enumerate}
\item Generalization Analysis for Deep Contrastive Representation Learning
\item Multiplex Dirichlet stochastic block model for clustering
  multidimensional compositional networks
\item BetaExplainer: A Probabilistic Method to Explain Graph Neural Networks
\item Bayesian Surrogate Training on Multiple Data Sources: A Hybrid Modeling
  Strategy
\item Scalable Temporal Anomaly Causality Discovery in Large Systems:
  Achieving Computational Efficiency with Binary Anomaly Flag Data
\item Conditional Diffusion Models Based Conditional Independence Testing
\item Generalized Bayesian deep reinforcement learning
\item A partial likelihood approach to tree-based density modeling and its
  application in Bayesian inference
\item A Mapper Algorithm with implicit intervals and its optimization
\item Learning Massive-scale Partial Correlation Networks in Clinical
  Multi-omics Studies with HP-ACCORD
\end{enumerate}

stat.TH
\begin{enumerate}
\item Optimality of the Right-Invariant Prior
\item The entropic optimal (self-)transport problem: Limit distributions for
  decreasing regularization with application to score function estimation
\item Causal Invariance Learning via Efficient Optimization of a Nonconvex
  Objective
\item A partial likelihood approach to tree-based density modeling and its
  application in Bayesian inference
\item Dual Unscented Kalman Filter Architecture for Sensor Fusion in Water
  Networks Leak Localization
\item Learning Massive-scale Partial Correlation Networks in Clinical
  Multi-omics Studies with HP-ACCORD
\item Well-Posedness and Stability of the Stochastic OGTT Model
\item Posterior asymptotics of high-dimensional spiked covariance model with
  inverse-Wishart prior
\item Model checking for high dimensional generalized linear models based on
  random projections
\item The Stein-log-Sobolev inequality and the exponential rate of convergence
  for the continuous Stein variational gradient descent method
\end{enumerate}

eess.IV
\begin{enumerate}
\item Are the Latent Representations of Foundation Models for Pathology
  Invariant to Rotation?
\item Towards Physically-Based Sky-Modeling
\item Ant Nest Detection Using Underground P-Band TomoSAR
\item Ensemble Learning and 3D Pix2Pix for Comprehensive Brain Tumor Analysis
  in Multimodal MRI
\item Point Cloud-Assisted Neural Image Compression
\item Flex-PE: Flexible and SIMD Multi-Precision Processing Element for AI
  Workloads
\item Fast-staged CNN Model for Accurate pulmonary diseases and Lung cancer
  detection
\item High-speed and High-quality Vision Reconstruction of Spike Camera with
  Spike Stability Theorem
\item Data-driven Precipitation Nowcasting Using Satellite Imagery
\item Block-Based Multi-Scale Image Rescaling
\end{enumerate}

eess.SP
\begin{enumerate}
\item Rate-Splitting Multiple Access for Integrated Sensing and
  Communications: A First Experimental Study
\item Soil moisture estimation of bare and vegetation-covered areas using a
  P/L/C-band SAR
\item Ant Nest Detection Using Underground P-Band TomoSAR
\item Scalable Data Transmission Framework for Earth Observation Satellites
  with Channel Adaptation
\item Sonar-based Deep Learning in Underwater Robotics: Overview, Robustness
  and Challenges
\item Evaluating the Efficacy of Vectocardiographic and ECG Parameters for
  Efficient Tertiary Cardiology Care Allocation Using Decision Tree Analysis
\item Acceleration and Parallelization Methods for ISRS EGN Model
\item On-the-Fly Interrogation of Mobile Passive Sensors from the Fusion of
  Optical and Radar Data
\item Capacity Analysis on OAM-Based Wireless Communications: An
  Electromagnetic Information Theory Perspective
\item Probabilistic GOSPA: A Metric for Performance Evaluation of Multi-Object
  Filters with Uncertainties
\end{enumerate}

econ.EM
\begin{enumerate}
\item Moderating the Mediation Bootstrap for Causal Inference
\item VAR models with an index structure: A survey with new results
\item Treatment Evaluation at the Intensive and Extensive Margins
\item Forecasting realized covariances using HAR-type models
\item Do LLMs Act as Repositories of Causal Knowledge?
\item An overview of meta-analytic methods for economic research
\item A Neyman-Orthogonalization Approach to the Incidental Parameter Problem
\item Geometric Deep Learning for Realized Covariance Matrix Forecasting
\item A Kernel Score Perspective on Forecast Disagreement and the Linear Pool
\item The Global Carbon Budget as a cointegrated system
\end{enumerate}

econ.GN
\begin{enumerate}
\item Multiplexing in Networks and Diffusion
\item Transition dynamics of electricity asset-owning firms
\item Binary or nonbinary? An evolutionary learning approach to gender
  identity
\item On Prior Confidence and Belief Updating
\item Strategically Acting on Information
\item Is Polarization an Inevitable Outcome of Similarity-Based Content
  Recommendations? -- Mathematical Proofs and Computational Validation
\item Re-examining the social impact of silver monetization in the Ming
  Dynasty from the perspective of supply and demand
\item Delving into Youth Perspectives on In-game Gambling-like Elements: A
  Proof-of-Concept Study Utilising Large Language Models for Analysing
  User-Generated Text Data
\item Does Low Spoilage Under Cold Conditions Foster Cultural Complexity
  During the Foraging Era? -- A Theoretical and Computational Inquiry
\item Emulating the Global Change Analysis Model with Deep Learning
\end{enumerate}

\end{small}

\subsection{Robustness Check: Q/A task with another embedding model}
\label[appendix]{sec:another-embedding-model}

\begin{figure}[h]
    \centering
    \includegraphics[width=0.7\linewidth]{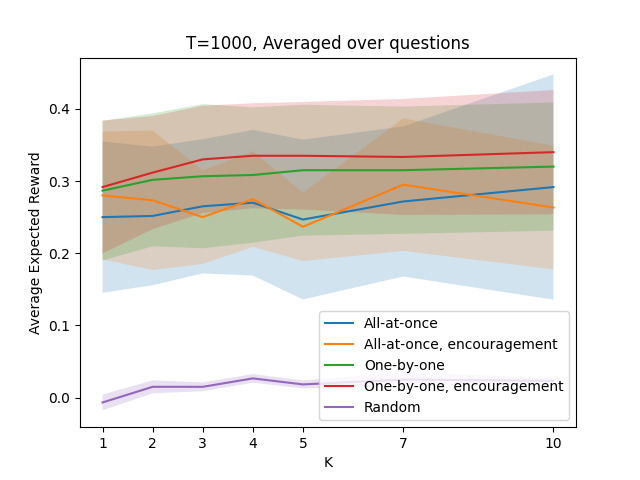}
    \caption{Results averaged over the first six questions, for embeddings generated using the universal sentence encoder.}
    \label{fig:universal}
\end{figure}

We ran a robustness check on the first six open-ended question experiments using the universal sentence encoder of~\citet{cer2018universal} as our embedding model.
Our results remain largely unchanged, and are summarized in~\Cref{fig:universal}.

\clearpage
\subsection{Additional Results: arXiv task}

\begin{figure}[h]
    \begin{subfigure}[b]{0.49\textwidth}
        \centering
        \includegraphics[width=\linewidth]{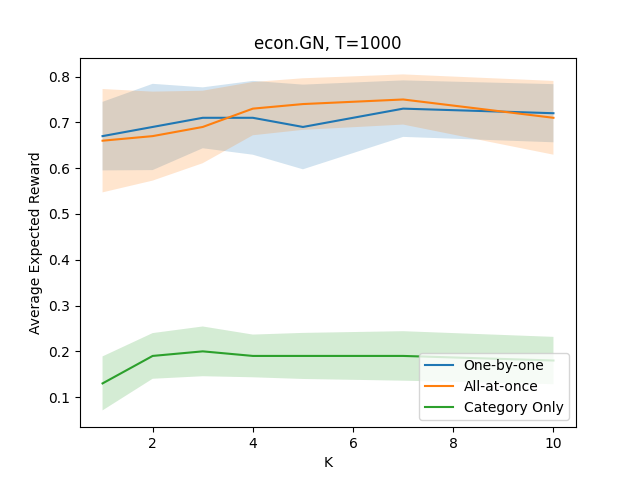}
    \end{subfigure}
    %
    \begin{subfigure}[b]{0.49\textwidth}
        \centering
        \includegraphics[width=\linewidth]{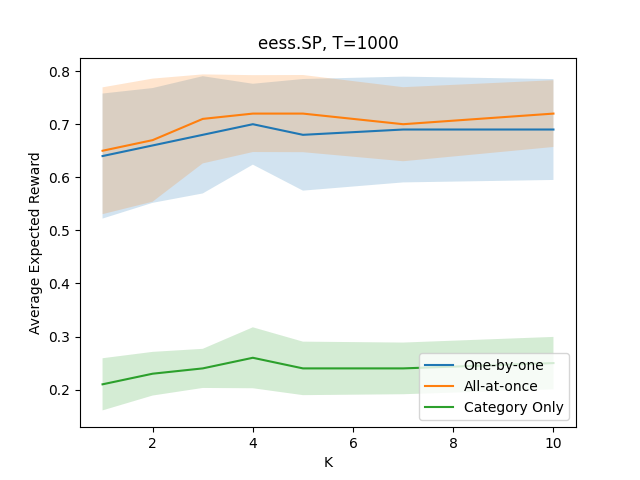}
    \end{subfigure}
    \begin{subfigure}[b]{0.49\textwidth}
        \centering
        \includegraphics[width=\linewidth]{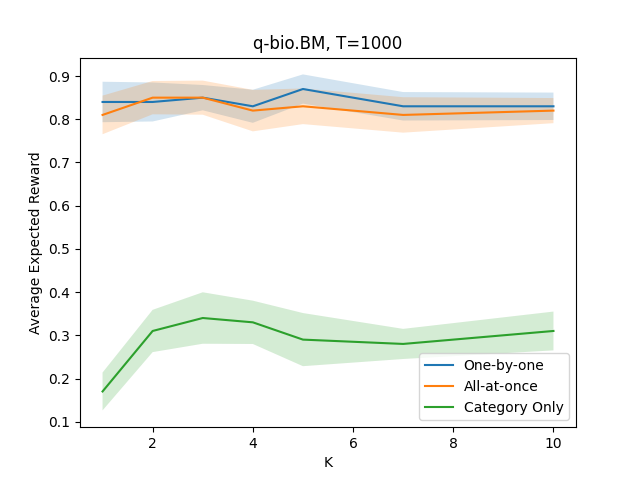}
    \end{subfigure}
    \begin{subfigure}[b]{0.49\textwidth}
        \centering
        \includegraphics[width=\linewidth]{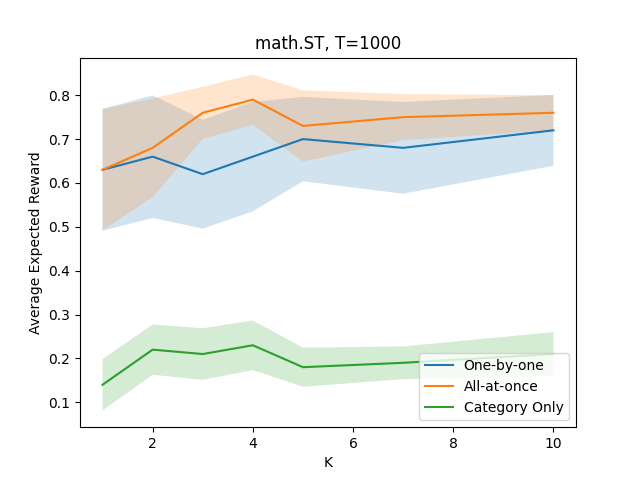}
    \end{subfigure}
    \caption{Results for other arXiv categories}
\end{figure}

\clearpage
\begin{table}[h]
\caption{Performance for all-at-once on arXiv tasks.}
\label{table-aao-arxiv}
\centering
\begin{tabular}{lcccc}
\toprule
 & K=1 & K=2 & K=5 & MMR ($K=5$)\\
\midrule
gr-qc & 0.63 & 0.64 & 0.68 & 0.40\\
hep-ex & 0.81 & 0.81 & 0.83 & 0.52\\
hep-lat & 0.72 & 0.72 & 0.72 & 0.43\\
hep-ph & 0.7 & 0.76 & 0.75 & 0.44\\
hep-th & 0.65 & 0.71 & 0.73 & 0.40\\
math-ph & 0.64 & 0.73 & 0.74 & 0.39\\
nucl-ex & 0.73 & 0.79 & 0.75 & 0.41\\
nucl-th & 0.65 & 0.69 & 0.71 & 0.41\\
quant-ph & 0.68 & 0.71 & 0.75 & 0.38\\
cs.AI & 0.66 & 0.71 & 0.72 & 0.55\\
cs.CL & 0.66 & 0.71 & 0.75 & 0.45\\
cs.CV & 0.72 & 0.74 & 0.71 & 0.47\\
cs.LG & 0.68 & 0.72 & 0.74 & 0.51\\
cs.NE & 0.71 & 0.78 & 0.78 & 0.40,\\
cs.RO & 0.76 & 0.79 & 0.78 & 0.51\\
cs.IT & 0.72 & 0.73 & 0.72 & 0.44\\
cs.CR & 0.7 & 0.72 & 0.74 & 0.51\\
cs.DS & 0.75 & 0.77 & 0.77 & 0.38\\
cs.HC & 0.75 & 0.75 & 0.75 & 0.52\\
math.AG & 0.7 & 0.78 & 0.78 & 0.43\\
math.AT & 0.68 & 0.7 & 0.71 & 0.39\\
math.AP & 0.7 & 0.79 & 0.78 & 0.40\\
math.CT & 0.65 & 0.69 & 0.71 & 0.38\\
math.GR & 0.73 & 0.77 & 0.76 & 0.38\\
math.NT & 0.73 & 0.79 & 0.77 & 0.38\\
math.OC & 0.77 & 0.79 & 0.76 & 0.44\\
math.ST & 0.63 & 0.68 & 0.73 & 0.44\\
q-bio.BM & 0.81 & 0.85 & 0.83 & 0.51\\
q-bio.GN & 0.76 & 0.78 & 0.79 & 0.49\\
q-bio.QM & 0.76 & 0.78 & 0.78 & 0.43\\
q-bio.PE & 0.8 & 0.82 & 0.8 & 0.53\\
q-fin.CP & 0.74 & 0.78 & 0.77 & 0.42\\
q-fin.PM & 0.74 & 0.77 & 0.78 & 0.44\\
q-fin.TR & 0.74 & 0.78 & 0.78 & 0.42\\
stat.AP & 0.73 & 0.69 & 0.75 & 0.40\\
stat.ML & 0.7 & 0.73 & 0.74 & 0.45\\
stat.TH & 0.65 & 0.67 & 0.78 & 0.44\\
eess.IV & 0.67 & 0.73 & 0.72 & 0.44\\
eess.SP & 0.65 & 0.67 & 0.72 & 0.39\\
econ.EM & 0.62 & 0.68 & 0.7 & 0.40\\
econ.GN & 0.66 & 0.67 & 0.74 & 0.38\\
\bottomrule
\end{tabular}
\end{table}

\begin{table}[h]
\caption{Performance for one-by-one on arXiv tasks.}
\label{table-obo-arxiv}
\centering
\begin{tabular}{lccc}
\toprule
 & K=1 & K=2 & K=5 \\
\midrule
gr-qc & 0.63 & 0.64 & 0.65\\
hep-ex & 0.78 & 0.76 & 0.81\\
hep-lat & 0.72 & 0.72 & 0.74\\
hep-ph & 0.7 & 0.72 & 0.73\\
hep-th & 0.64 & 0.69 & 0.68\\
math-ph & 0.65 & 0.71 & 0.71\\
nucl-ex & 0.72 & 0.74 & 0.76\\
nucl-th & 0.64 & 0.67 & 0.71\\
quant-ph & 0.7 & 0.71 & 0.71\\
cs.AI & 0.7 & 0.74 & 0.73\\
cs.CL & 0.69 & 0.71 & 0.75\\
cs.CV & 0.73 & 0.74 & 0.77\\
cs.LG & 0.67 & 0.72 & 0.72\\
cs.NE & 0.74 & 0.76 & 0.77\\
cs.RO & 0.78 & 0.79 & 0.78\\
cs.IT & 0.76 & 0.75 & 0.75\\
cs.CR & 0.72 & 0.71 & 0.74\\
cs.DS & 0.75 & 0.78 & 0.78\\
cs.HC & 0.72 & 0.72 & 0.73\\
math.AG & 0.68 & 0.77 & 0.78\\
math.AT & 0.63 & 0.69 & 0.7\\
math.AP & 0.72 & 0.75 & 0.76\\
math.CT & 0.62 & 0.71 & 0.73\\
math.GR & 0.71 & 0.76 & 0.74\\
math.NT & 0.73 & 0.75 & 0.73\\
math.OC & 0.73 & 0.77 & 0.8\\
math.ST & 0.63 & 0.66 & 0.7\\
q-bio.BM & 0.84 & 0.84 & 0.87\\
q-bio.GN & 0.69 & 0.76 & 0.75\\
q-bio.QM & 0.76 & 0.8 & 0.77\\
q-bio.PE & 0.8 & 0.79 & 0.82\\
q-fin.CP & 0.71 & 0.72 & 0.76\\
q-fin.PM & 0.67 & 0.77 & 0.73\\
q-fin.TR & 0.72 & 0.74 & 0.75\\
stat.AP & 0.73 & 0.72 & 0.79\\
stat.ML & 0.74 & 0.75 & 0.77\\
stat.TH & 0.61 & 0.64 & 0.72\\
eess.IV & 0.71 & 0.72 & 0.74\\
eess.SP & 0.64 & 0.66 & 0.68\\
econ.EM & 0.66 & 0.66 & 0.67\\
econ.GN & 0.67 & 0.69 & 0.69\\
\bottomrule
\end{tabular}
\end{table}

\begin{table}[h]
\caption{Performance for Category Only baseline on arXiv tasks.}
\label{table-co-arxiv}
\centering
\begin{tabular}{lccc}
\toprule
 & K=1 & K=2 & K=5 \\
\midrule
gr-qc & 0.28 & 0.32 & 0.36\\
hep-ex & 0.25 & 0.35 & 0.38\\
hep-lat & 0.31 & 0.31 & 0.38\\
hep-ph & 0.25 & 0.26 & 0.32\\
hep-th & 0.24 & 0.26 & 0.27\\
math-ph & 0.22 & 0.3 & 0.3\\
nucl-ex & 0.37 & 0.38 & 0.35\\
nucl-th & 0.31 & 0.32 & 0.33\\
quant-ph & 0.23 & 0.27 & 0.27\\
cs.AI & 0.15 & 0.17 & 0.18\\
cs.CL & 0.14 & 0.19 & 0.26\\
cs.CV & 0.19 & 0.22 & 0.32\\
cs.LG & 0.21 & 0.19 & 0.24\\
cs.NE & 0.27 & 0.31 & 0.3\\
cs.RO & 0.27 & 0.27 & 0.31\\
cs.IT & 0.29 & 0.31 & 0.34\\
cs.CR & 0.21 & 0.24 & 0.32\\
cs.DS & 0.2 & 0.2 & 0.22\\
cs.HC & 0.12 & 0.2 & 0.23\\
math.AG & 0.32 & 0.33 & 0.33\\
math.AT & 0.33 & 0.34 & 0.4\\
math.AP & 0.19 & 0.23 & 0.32\\
math.CT & 0.25 & 0.23 & 0.29\\
math.GR & 0.25 & 0.28 & 0.34\\
math.NT & 0.21 & 0.29 & 0.32\\
math.OC & 0.2 & 0.27 & 0.3\\
math.ST & 0.14 & 0.22 & 0.18\\
q-bio.BM & 0.17 & 0.31 & 0.29\\
q-bio.GN & 0.2 & 0.25 & 0.32\\
q-bio.QM & 0.03 & 0.12 & 0.1\\
q-bio.PE & 0.28 & 0.32 & 0.31\\
q-fin.CP & 0.34 & 0.36 & 0.37\\
q-fin.PM & 0.39 & 0.39 & 0.43\\
q-fin.TR & 0.33 & 0.34 & 0.38\\
stat.AP & 0.07 & 0.12 & 0.1\\
stat.ML & 0.19 & 0.21 & 0.28\\
stat.TH & -0.01 & 0.07 & 0.12\\
eess.IV & 0.18 & 0.19 & 0.24\\
eess.SP & 0.21 & 0.23 & 0.24\\
econ.EM & 0.21 & 0.32 & 0.37\\
econ.GN & 0.13 & 0.19 & 0.19\\
\bottomrule
\end{tabular}
\end{table}

\clearpage
\subsection{Additional Results: QA task}
\label[appendix]{app:QA-additional}

While previously we presented the plot for Question 0 (in \Cref{fig:ucb}), below are the individual plots for the remaining $9$ questions, in the same notation. Throughout, we used the Sentence-BERT encoder.

\begin{center}
\includegraphics[width=0.55\linewidth]{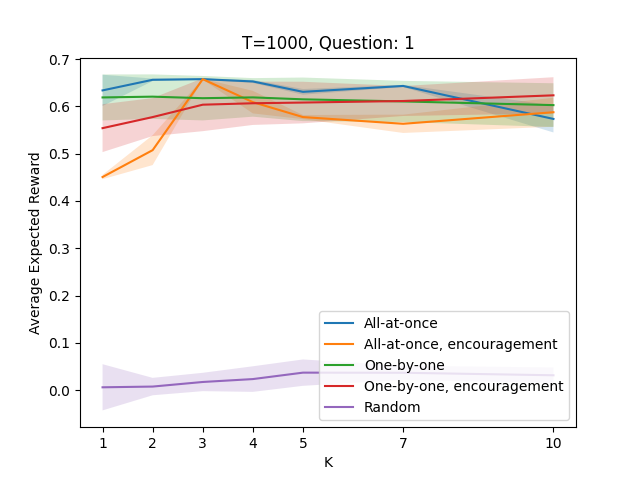}
\includegraphics[width=0.55\linewidth]{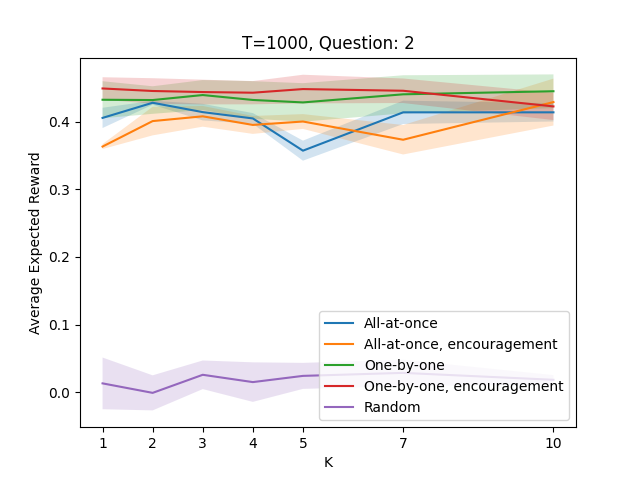}
\includegraphics[width=0.55\linewidth]{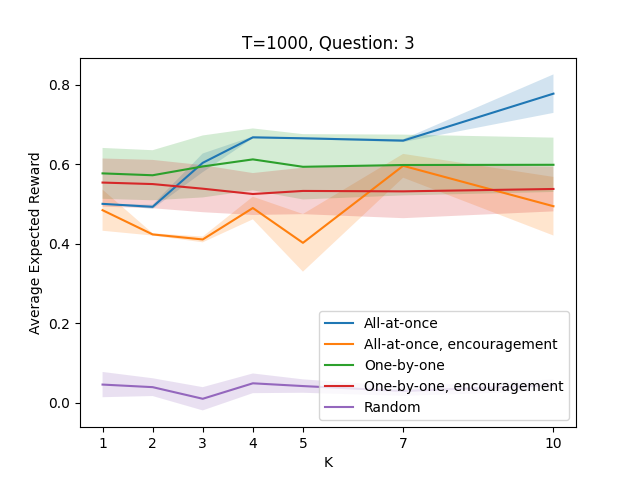}
\includegraphics[width=0.55\linewidth]{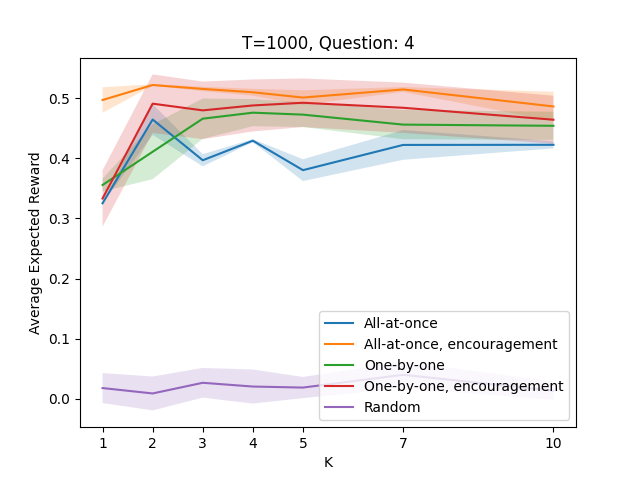}
\includegraphics[width=0.55\linewidth]{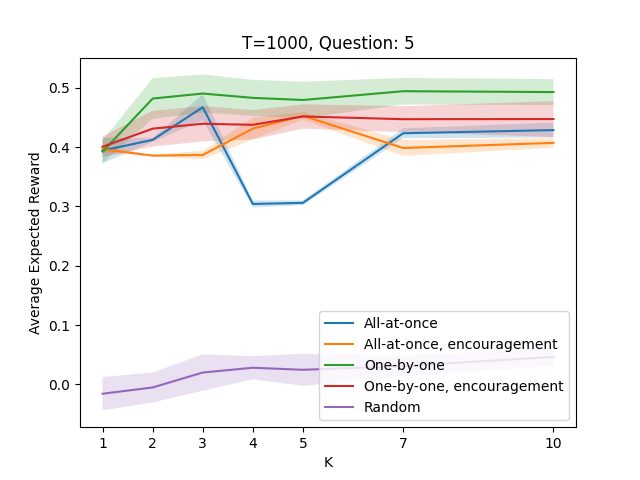}
\includegraphics[width=0.55\linewidth]{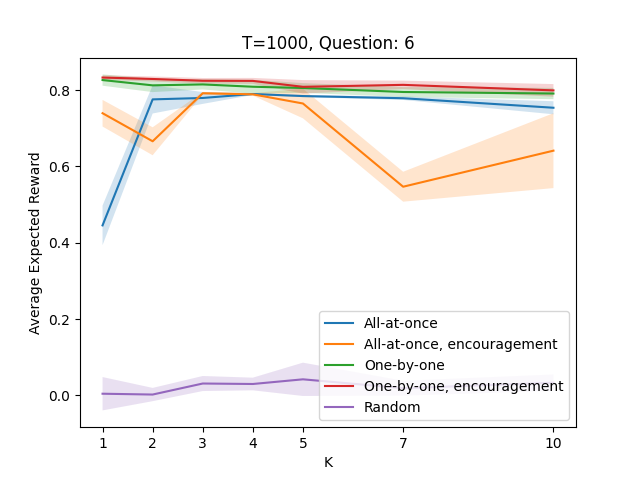}
\includegraphics[width=0.55\linewidth]{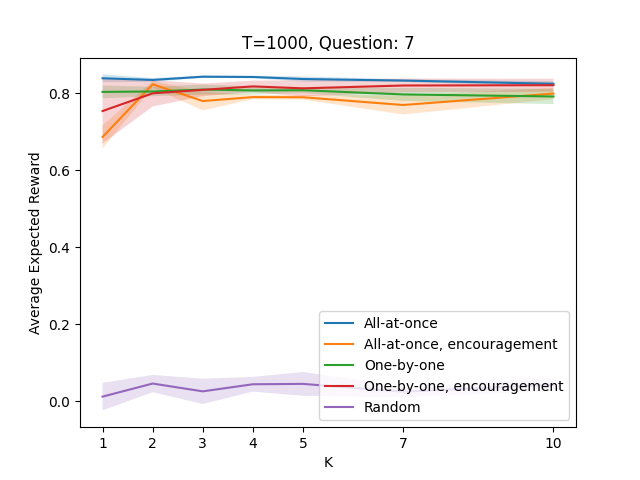}
\includegraphics[width=0.55\linewidth]{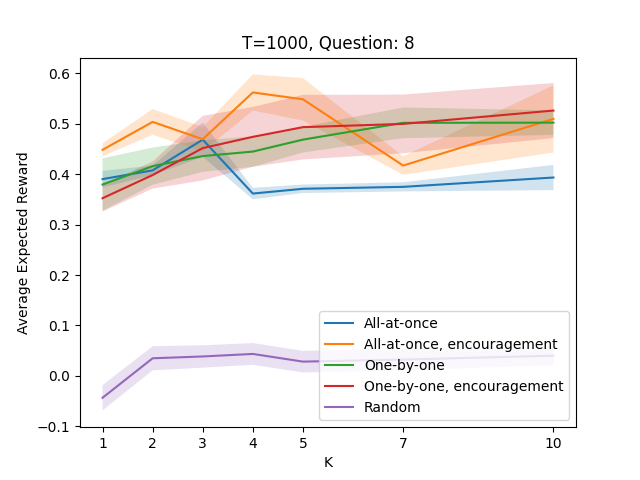}
\includegraphics[width=0.55\linewidth]{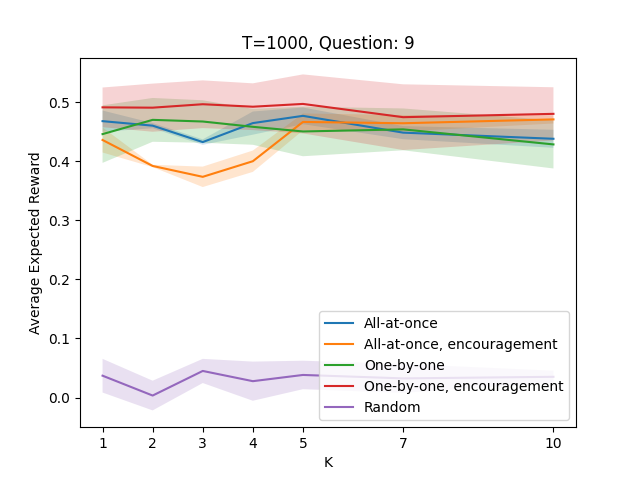}
\end{center}










\clearpage

\begin{table}[t]
\caption{Performance comparison for all-at-once on open-ended questions.}
\label{table-i1}
\centering
\begin{tabular}{lccccccc}
\toprule
 & K=1 & K=2 & K=3 & K=4 & K=5 & K=7 & K=10 \\
\midrule
Q0: & 0.39 & 0.4 & 0.4 & 0.4 & 0.41 & 0.42 & 0.56 \\
Q1: & 0.63 & 0.66 & 0.66 & 0.65 & 0.63 & 0.64 & 0.57 \\
Q2: & 0.41 & 0.43 & 0.41 & 0.4 & 0.36 & 0.41 & 0.41 \\
Q3: & 0.5 & 0.49 & 0.6 & 0.67 & 0.67 & 0.66 & 0.78 \\
Q4: & 0.33 & 0.46 & 0.4 & 0.43 & 0.38 & 0.42 & 0.42 \\
Q5: & 0.39 & 0.41 & 0.47 & 0.3 & 0.31 & 0.42 & 0.43 \\
Q6: & 0.45 & 0.78 & 0.78 & 0.79 & 0.78 & 0.78 & 0.75 \\
Q7: & 0.84 & 0.84 & 0.84 & 0.84 & 0.84 & 0.83 & 0.83 \\
Q8: & 0.39 & 0.41 & 0.47 & 0.36 & 0.37 & 0.37 & 0.39 \\
Q9: & 0.47 & 0.46 & 0.43 & 0.46 & 0.48 & 0.45 & 0.44 \\
\bottomrule
\end{tabular}
\end{table}

\begin{table}[t]
\caption{Performance comparison for all-at-once with encouragement  on open-ended questions.}
\label{table-i2}
\centering
\begin{tabular}{lccccccc}
\toprule
 & K=1 & K=2 & K=3 & K=4 & K=5 & K=7 & K=10 \\
\midrule
Q0: & 0.4 & 0.4 & 0.34 & 0.35 & 0.37 & 0.52 & 0.55 \\
Q1: & 0.45 & 0.51 & 0.66 & 0.61 & 0.58 & 0.56 & 0.59 \\
Q2: & 0.36 & 0.4 & 0.41 & 0.4 & 0.4 & 0.37 & 0.43 \\
Q3: & 0.48 & 0.42 & 0.41 & 0.49 & 0.4 & 0.6 & 0.49 \\
Q4: & 0.5 & 0.52 & 0.52 & 0.51 & 0.5 & 0.51 & 0.49 \\
Q5: & 0.4 & 0.39 & 0.39 & 0.43 & 0.45 & 0.4 & 0.41 \\
Q6: & 0.74 & 0.67 & 0.79 & 0.79 & 0.77 & 0.55 & 0.64 \\
Q7: & 0.69 & 0.82 & 0.78 & 0.79 & 0.79 & 0.77 & 0.8 \\
Q8: & 0.45 & 0.5 & 0.47 & 0.56 & 0.55 & 0.42 & 0.51 \\
Q9: & 0.44 & 0.39 & 0.37 & 0.4 & 0.47 & 0.46 & 0.47 \\
\bottomrule
\end{tabular}
\end{table}

\begin{table}[t]
\caption{Performance comparison for one-by-one  on open-ended questions.}
\label{table-i3}
\centering
\begin{tabular}{lccccccc}
\toprule
 & K=1 & K=2 & K=3 & K=4 & K=5 & K=7 & K=10 \\
\midrule
Q0: & 0.33 & 0.4 & 0.4 & 0.43 & 0.43 & 0.47 & 0.52 \\
Q1: & 0.62 & 0.62 & 0.62 & 0.62 & 0.61 & 0.61 & 0.6 \\
Q2: & 0.43 & 0.43 & 0.44 & 0.43 & 0.43 & 0.44 & 0.45 \\
Q3: & 0.58 & 0.57 & 0.59 & 0.61 & 0.59 & 0.6 & 0.6 \\
Q4: & 0.36 & 0.41 & 0.47 & 0.48 & 0.47 & 0.46 & 0.45 \\
Q5: & 0.39 & 0.48 & 0.49 & 0.48 & 0.48 & 0.49 & 0.49 \\
Q6: & 0.83 & 0.81 & 0.82 & 0.81 & 0.81 & 0.8 & 0.79 \\
Q7: & 0.8 & 0.81 & 0.81 & 0.81 & 0.81 & 0.8 & 0.79 \\
Q8: & 0.38 & 0.42 & 0.44 & 0.44 & 0.47 & 0.5 & 0.5 \\
Q9: & 0.45 & 0.47 & 0.47 & 0.46 & 0.45 & 0.45 & 0.43 \\
\bottomrule
\end{tabular}
\end{table}

\begin{table}[t]
\caption{Performance comparison for one-by-one with encouragement on open-ended questions.}
\label{table-i4}
\centering
\begin{tabular}{lccccccc}
\toprule
 & K=1 & K=2 & K=3 & K=4 & K=5 & K=7 & K=10 \\
\midrule
Q0: & 0.35 & 0.37 & 0.5 & 0.54 & 0.55 & 0.58 & 0.59 \\
Q1: & 0.55 & 0.58 & 0.6 & 0.61 & 0.61 & 0.61 & 0.62 \\
Q2: & 0.45 & 0.45 & 0.44 & 0.44 & 0.45 & 0.45 & 0.42 \\
Q3: & 0.55 & 0.55 & 0.54 & 0.53 & 0.53 & 0.53 & 0.54 \\
Q4: & 0.33 & 0.49 & 0.48 & 0.49 & 0.49 & 0.48 & 0.46 \\
Q5: & 0.4 & 0.43 & 0.44 & 0.44 & 0.45 & 0.45 & 0.45 \\
Q6: & 0.83 & 0.83 & 0.82 & 0.82 & 0.81 & 0.81 & 0.8 \\
Q7: & 0.75 & 0.8 & 0.81 & 0.82 & 0.81 & 0.82 & 0.82 \\
Q8: & 0.35 & 0.4 & 0.45 & 0.47 & 0.49 & 0.5 & 0.53 \\
Q9: & 0.49 & 0.49 & 0.5 & 0.49 & 0.5 & 0.47 & 0.48 \\
\bottomrule
\end{tabular}
\end{table}

\begin{table}[t]
\caption{Performance comparison for random actions on open-ended questions.}
\label{table-i5}
\centering
\begin{tabular}{lccccccc}
\toprule
 & K=1 & K=2 & K=3 & K=4 & K=5 & K=7 & K=10 \\
\midrule
Q0: & -0.01 & -0.0 & 0.01 & 0.01 & 0.03 & 0.03 & 0.02 \\
Q1: & 0.01 & 0.01 & 0.02 & 0.02 & 0.04 & 0.04 & 0.03 \\
Q2: & 0.01 & -0.0 & 0.03 & 0.02 & 0.02 & 0.03 & 0.02 \\
Q3: & 0.05 & 0.04 & 0.01 & 0.05 & 0.04 & 0.03 & 0.04 \\
Q4: & 0.02 & 0.01 & 0.03 & 0.02 & 0.02 & 0.04 & 0.01 \\
Q5: & -0.02 & -0.01 & 0.02 & 0.03 & 0.02 & 0.03 & 0.05 \\
Q6: & 0.0 & 0.0 & 0.03 & 0.03 & 0.04 & 0.02 & 0.03 \\
Q7: & 0.01 & 0.05 & 0.03 & 0.04 & 0.04 & 0.02 & 0.04 \\
Q8: & -0.04 & 0.03 & 0.04 & 0.04 & 0.03 & 0.03 & 0.04 \\
Q9: & 0.04 & 0.0 & 0.04 & 0.03 & 0.04 & 0.03 & 0.04 \\
\bottomrule
\end{tabular}
\end{table}


\clearpage
\subsection{Robustness Check with Other Models}\label{sec:explore-robustness}

We include results for Qwen2.5-7B-Instruct ($\qwen$), Gemma 3 12B ($\gemma$), and Mistral-7B-Instruct-v0.3 ($\mistral$) on our explore Q/A puzzles, all in the setting of~\Cref{fig:ucb}.
As was the case with the GPT models, performance increased with $K$ when the candidate answers are generated one-by-one.
When generating answers all-at-once, we observed that $\qwen$ and $\mistral$ had trouble generating concise answers, leading to a drop in performance as $K$ increases.

\begin{center}
    \includegraphics[width=0.5\linewidth]{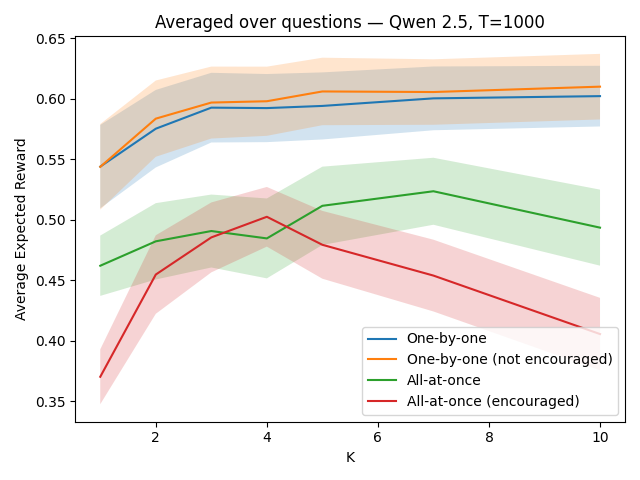}
    \includegraphics[width=0.5\linewidth]{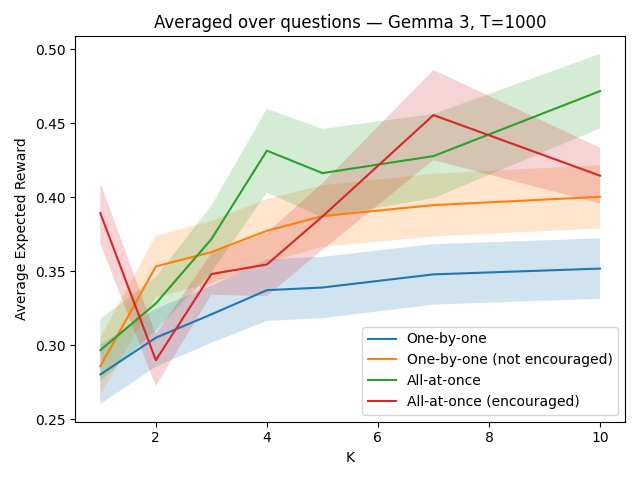}
    \includegraphics[width=0.5\linewidth]{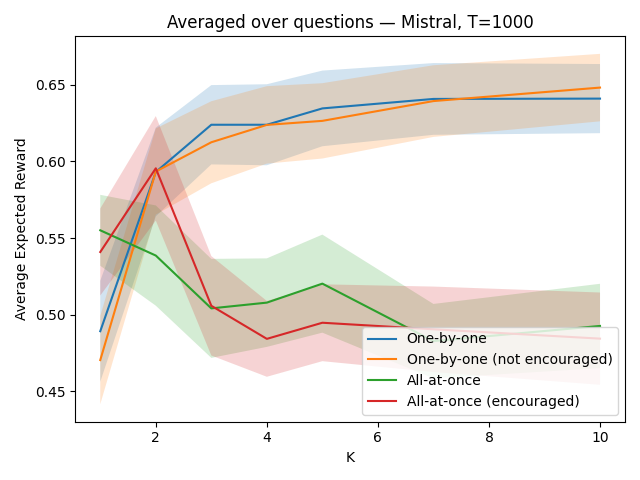}
\end{center}

\clearpage




\clearpage
\subsection{Benchmarking encoders}\label{sec:encoders}

Here we benchmark the two encoders we use (Sentence-BERT and the universal sentence encoder) by measuring the cosine similarity between semantically similar/different words.

\begin{table}[h]
\caption{Cosine similarity of different words.}
\label{tab:embedding}
\centering
\begin{tabular}{lcc}
\toprule
 & Sentence-BERT & Universal Sentence Encoder \\
\midrule
dog, tacos: & 0.25 & 0.24\\
Pittsburgh, tiki bar: & 0.12 & 0.17\\
Honolulu, tiki bar: & 0.30 & 0.25\\
Pittsburgh, Honolulu: & 0.41 & 0.29\\
angel, devil: & 0.48 & 0.54\\
machine learning, artificial intelligence: & 0.70 & 0.58\\
war, peace: & 0.61 & 0.49\\
love, hate: & 0.49 & 0.59\\
love, affection: & 0.62 & 0.56\\
war, battle: & 0.74 & 0.57\\
machine learning, battle: & 0.25 & 0.19\\
\bottomrule
\end{tabular}
\end{table}

The similarity scores of both models in~\Cref{tab:embedding} suggest that while the embeddings produced by both embedding models are generally ``in the ballpark'' of what one would consider ``similar''/``different'', they are still a somewhat coarse measure of distance, which may explain the similar performance of our different prompting strategies.

\subsection{Explore Experiments on the Movie Lens Dataset}
\label[appendix]{app:SVD-imputation}

To test the sensitivity to Soft-Impute, we repeated our Movie Lens experiments using a truncated SVD imputation, which is a popular alternative.
Our results were largely unchanged (Table~\ref{tab:movielens2}).

\begin{table}
\centering
\caption{Average Reward for $\qwen$, $\mistral$, and Baselines for $K \in \{10, 19\}$ in the MovieLens Task using truncated SVD imputation.}
\begin{tabular}{l r r}
\hline
 $\AveRew$ &  $K=10$ & $K=19$ \\
\hline
$\qwen$             & $613.80$ & $609.80$ \\
$\mistral$          & $642.29$ & $604.53$ \\
Random & $550.99$ & $563.93$ \\
Genre-based      & ---    & $573.90$ \\
\hline
\end{tabular}
\label{tab:movielens2}
\end{table}

Prompt:

\begin{quote}
    [SYSTEM] You are a movie expert helping a user choose a movie.

    [USER] Here is a list of movies with their numeric IDs:\{movie\_list\_str\}

    From this list, choose {K} movies. You don't know what taste in movies the user has, so select a diverse set of movies from different genres such that they will most likely enjoy at least one of the movies you select. Respond ONLY with the {K} numeric IDs, one per line, with no extra text. Do NOT consider a movie's popularity when deciding whether to select it.
\end{quote}

\clearpage

\section{Cost and Latency Estimates}

In this section, we provide rough cost and latency estimates for our experiments.
We highlight \gptfivetwo and \gptfouro, as they appeared most frequently in our main body experiments.
Latencies for other models were similar, and costs for the other models we used may be found on the OpenAI platform and HuggingFace.

Each exploit call processes the full history in-context, leading to prompt sizes that scale linearly with the number of rounds.
In our experiments, this ranges from a few hundred tokens (small MAB instances) to several thousand tokens per call (e.g., $5000-8000$ tokens for the Figure 2 numerical CB setting with $T=100$). Larger settings (e.g., $T=4000$) exceed context limits and require summarization, reducing effective prompt sizes to $1000-8000$ tokens.

Regarding costs (in USD), \gptfivetwo is currently priced at $\$1.25$ per 1M input tokens and $\$10.00$ per 1M output tokens.
\gptfouro is priced at $\$2.50$ per 1M input tokens and $\$10.00$ per 1M output tokens.
Using these prices, a single ``run;; of each experiment typically costs in the $\$0.5-\$5.00$ range for the main numerical settings.

Regarding latency, accessing \gptfivetwo and \gptfouro through the OpenAI API results in $60$ and $51$ tokens-per-second (respectively; according to OpenRouter).
Converting these throughputs to latency, this corresponds to roughly $17$ ms/token (\gptfivetwo) and $20$ ms/token (\gptfouro).
For typical exploit calls with $5000-8000$ tokens, this yields an estimated $80-230$ seconds per call depending on the model.
Smaller MAB instances require only $10-60$ seconds per call, while summarized large-CB instances again fall in the $20-200$ second range.

Aggregating over an experiment with approximately $100$ exploit calls, this implies total wall-clock times about $30$ minutes to several hours for LLM-only pipelines, even without tool use.
This is consistent with our measured runtimes (minutes without tools, hours with tools), and contrasts with regression-based solvers, which run in milliseconds to seconds for the same tasks.

These estimates are for LLM-only calls.
Tool augmentation is much more expensive in practice: in our Figure~\ref{fig:numerical} experiment, the code interpreter increased costs to about $\$70$ and latency to $6$+ hours.

Our exploration experiments were significantly less expensive to run in terms of both cost and latency, as only one API call was required to generate the action discretization. No other significant compute resources were used, i.e. our regression-based solvers were implemented on a laptop. 

%% file: bib-AGT.bib
@inproceedings{KleinbergL03,
  author    = {Robert D. Kleinberg and
               Frank T. Leighton},
  title     = {The Value of Knowing a Demand Curve: Bounds on Regret for
               Online Posted-Price Auctions},
  booktitle = FOCS,
  publisher = IEEEpub,
  pages     = {594-605},
  year      = {2003},
}


%% file: bib-LLMs.bib
@inproceedings{xu2022prompting,
  title={Prompting decision transformer for few-shot policy generalization},
  author={Xu, Mengdi and Shen, Yikang and Zhang, Shun and Lu, Yuchen and Zhao, Ding and Tenenbaum, Joshua and Gan, Chuang},
  booktitle={international conference on machine learning},
  pages={24631--24645},
  year={2022},
  organization={PMLR}
}

@article{wang2023voyager,
  title={Voyager: An open-ended embodied agent with large language models},
  author={Wang, Guanzhi and Xie, Yuqi and Jiang, Yunfan and Mandlekar, Ajay and Xiao, Chaowei and Zhu, Yuke and Fan, Linxi and Anandkumar, Anima},
  journal={arXiv:2305.16291},
  year={2023}
}

@article{lin2023transformers,
  title={Transformers as Decision Makers: Provable In-Context Reinforcement Learning via Supervised Pretraining},
  author={Lin, Licong and Bai, Yu and Mei, Song},
  journal={arXiv preprint arXiv:2310.08566},
  year={2023}
}

@article{laskin2022context,
  title={In-context reinforcement learning with algorithm distillation},
  author={Laskin, Michael and Wang, Luyu and Oh, Junhyuk and Parisotto, Emilio and Spencer, Stephen and Steigerwald, Richie and Strouse, DJ and Hansen, Steven and Filos, Angelos and Brooks, Ethan and others},
  journal={arXiv preprint arXiv:2210.14215},
  year={2022}
}

@article{raparthy2023generalization,
  title={Generalization to New Sequential Decision Making Tasks with In-Context Learning},
  author={Raparthy, Sharath Chandra and Hambro, Eric and Kirk, Robert and Henaff, Mikael and Raileanu, Roberta},
  journal={arXiv preprint arXiv:2312.03801},
  year={2023}
}


%% file: bib-RL.bib
@misc{RLTheoryBook-20,
  title={Reinforcement learning: Theory and algorithms},
  author={Agarwal, Alekh and Jiang, Nan and Kakade, Sham M and Sun, Wen},
  note={Book draft, circulated since 2019. Available at \texttt{https://rltheorybook.github.io}.},
  year={2020}
}


%% file: bib-abbrv.bib
@String{ proc         =  ""}

@String{ FOCS         = " IEEE Symp. on Foundations of Computer Science (FOCS)"}

@String{ STOC         = " ACM Symp. on Theory of Computing (STOC)"}

@String{ SODA         = " ACM-SIAM Symp. on Discrete Algorithms (SODA)"}

@String{ NIPS         = " Advances in Neural Information Processing Systems (NIPS)"}

@String{ NeurIPS         = " Advances in Neural Information Processing Systems (NeurIPS)"}

@String{ COLT         = " Conf. on Learning Theory (COLT)"}

@String{ AISTATS = " Intl. Conf. on Artificial Intelligence and Statistics (AISTATS)"}

@String{ WWW  = " Intl. World Wide Web Conf. (WWW)"}

@String{ SIGIR        = " ACM Intl. Conf. on Research and Development in Information Retrieval (SIGIR)"}

@String{ KDD  = " ACM SIGKDD Intl. Conf. on Knowledge Discovery and Data Mining (KDD)"}

@String{ JACM   = "J. of the ACM"}

@String{ JMLR         = "J. of Machine Learning Research (JMLR)"}

@String{ CHI  = " Conf. on Human Factors in Computing Systems (CHI)"}

@String{ IEEEpub = ""}


%% file: bib-bandits.bib
@inproceedings{Reyzin-aistats11-linear,
   author =         "Wei Chu and Lihong Li and Lev Reyzin and Robert E. Schapire",
   title =         "{Contextual Bandits with Linear Payoff Functions}",
   booktitle         = proc # "14th" # AISTATS,
   year         = "2011"
}

@inproceedings{Csaba-nips11,
  author    = {Yasin Abbasi-Yadkori and D{\'a}vid P{\'a}l and Csaba Szepesv{\'a}ri},
  title     = {Improved Algorithms for Linear Stochastic Bandits},
  booktitle = proc # "25th" # NIPS,
  year      = {2011},
  pages     = {2312-2320}
}

@inproceedings{beygelzimer2009offset,
  title={The offset tree for learning with partial labels},
  author={Beygelzimer, Alina and Langford, John},
  booktitle= proc # "15th" # KDD,
  year={2009}
}

@inproceedings{Langford-www10,
   author =         "Lihong Li and Wei Chu and John Langford and Robert E. Schapire",
   title =         "{A contextual-bandit approach to personalized news article recommendation}",
   booktitle         = proc # "19th" # WWW,
   pagesX         = {661--670},
   year         = "2010"
}

@article{DR-StatScience14,
  author    = {Miroslav Dud{\'{\i}}k and
               Dumitru Erhan and
               John Langford and
               Lihong Li},
  title     = {Doubly Robust Policy Evaluation and Optimization},
  journal = {Statistical Science},
  volume    = {29},
  number    = {4},
  pages     = {1097--1104},
  year      = {2014}
}

@article{bandits-ucb1,
  author    = {Peter Auer and Nicol{\`o} Cesa-Bianchi and Paul Fischer},
  title     = {Finite-time Analysis of the Multiarmed Bandit Problem.},
  journal   = {Machine Learning},
  volume    = {47},
  number    = {2-3},
  year      = {2002},
  pages     = {235-256}
}

@article{xbandits-nips08,
   author =         "S\'{e}bastien Bubeck and R\'{e}mi Munos and Gilles Stoltz and Csaba Szepesvari",
   title =         "{Online Optimization in X-Armed Bandits}",
   journal         = JMLR,
   volume         = "12",
   pages         = {1587--1627},
   year     = "2011",
   note = "Preliminary version in \emph{NIPS 2008}."
}


%% file: bib-slivkins.bib
@article{slivkins-MABbook,
    author = "Aleksandrs Slivkins",
    title= "Introduction to Multi-Armed Bandits",
    journal = "Foundations and Trends$\circledR$ in Machine Learning",
    volume = {12},
    number = {1-2},
    pages = {1-286},
    month = nov,
    year = "2019",
    note = "Published with \emph{Now Publishers} (Boston, MA, USA). Also available at
        {\tt https://arxiv.org/abs/1904.07272}."
}

@inproceedings{LipschitzMAB-stoc08,
   author =  "Robert Kleinberg and Aleksandrs Slivkins and Eli Upfal",
   title =         "Multi-Armed Bandits in Metric Spaces",
   booktitle         = proc # "40th" # STOC,
   pages         = {681-690},
   year         = "2008"
}

@article{LipschitzMAB-JACM,
   author =  "Robert Kleinberg and Aleksandrs Slivkins and Eli Upfal",
   title =         "Bandits and Experts in Metric Spaces",
   journal = JACM,
   volume = "66",
    number = "4",
    pages     = {30:1--30:77},
    month = may,
   year      = "2019",
   note     = "Merged and revised version of conference papers
    in {\em ACM STOC 2008} and {\em ACM-SIAM SODA 2010}.
    Also available at {\tt http://arxiv.org/abs/1312.1277}"
}

@article{contextualMAB-colt11,
  author    = {Aleksandrs Slivkins},
  title     = {Contextual bandits with similarity information},
  journal   = JMLR,
  volume    = {15},
  number    = {1},
  pages     = {2533--2568},
  year      = {2014},
  note      = "Preliminary version in \emph{COLT 2011}."
 }


%% file: refs.bib
@article{coda2024cogbench,
  title={CogBench: a large language model walks into a psychology lab},
  author={Coda-Forno, Julian and Binz, Marcel and Wang, Jane X and Schulz, Eric},
  journal={arXiv:2402.18225},
  year={2024}
}

@article{hayes2024relative,
  title={Relative Value Biases in Large Language Models},
  author={Hayes, William M and Yax, Nicolas and Palminteri, Stefano},
  journal={arXiv:2401.14530},
  year={2024}
}

@article{schubert2024context,
  title={In-context learning agents are asymmetric belief updaters},
  author={Schubert, Johannes A and Jagadish, Akshay K and Binz, Marcel and Schulz, Eric},
  journal={arXiv:2402.03969},
  year={2024}
}

@inproceedings{wu2024smartplay,
title={SmartPlay: {A} Benchmark for {LLM}s as Intelligent Agents},
author={Yue Wu and Xuan Tang and Tom Mitchell and Yuanzhi Li},
booktitle={International Conference on Learning Representations},
year={2024},
}

@article{lehnert2024beyond,
  title={Beyond A*: Better Planning with Transformers via Search Dynamics Bootstrapping},
  author={Lehnert, Lucas and Sukhbaatar, Sainbayar and Mcvay, Paul and Rabbat, Michael and Tian, Yuandong},
  journal={arXiv preprint arXiv:2402.14083},
  year={2024}
}

@article{li2024stride,
  title={STRIDE: A Tool-Assisted LLM Agent Framework for Strategic and Interactive Decision-Making},
  author={Li, Chuanhao and Yang, Runhan and Li, Tiankai and Bafarassat, Milad and Sharifi, Kourosh and Bergemann, Dirk and Yang, Zhuoran},
  journal={arXiv preprint arXiv:2405.16376},
  year={2024}
}

@article{hao2023reasoning,
  title={Reasoning with language model is planning with world model},
  author={Hao, Shibo and Gu, Yi and Ma, Haodi and Hong, Joshua Jiahua and Wang, Zhen and Wang, Daisy Zhe and Hu, Zhiting},
  journal={arXiv preprint arXiv:2305.14992},
  year={2023}
}

@article{zhou2023language,
  title={Language agent tree search unifies reasoning acting and planning in language models},
  author={Zhou, Andy and Yan, Kai and Shlapentokh-Rothman, Michal and Wang, Haohan and Wang, Yu-Xiong},
  journal={arXiv preprint arXiv:2310.04406},
  year={2023}
}

@article{zhao2024large,
  title={Large language models as commonsense knowledge for large-scale task planning},
  author={Zhao, Zirui and Lee, Wee Sun and Hsu, David},
  journal={Advances in Neural Information Processing Systems},
  volume={36},
  year={2024}
}

@article{black2024pi_0,
  title={A Vision-Language-Action Flow Model for General Robot Control},
  author={Black, Kevin and Brown, Noah and Driess, Danny and Esmail, Adnan and Equi, Michael and Finn, Chelsea and Fusai, Niccolo and Groom, Lachy and Hausman, Karol and Ichter, Brian and others},
  journal={arXiv preprint arXiv:2410.24164},
  year={2024}
}

@inproceedings{krishnamurthy2024can,
  author       = {Akshay Krishnamurthy and
                  Keegan Harris and
                  Dylan J. Foster and
                  Cyril Zhang and
                  Aleksandrs Slivkins},
  title        = {Can large language models explore in-context?},
  booktitle     = "NeurIPS",
  year         = {2024}
}

@article{nie2024evolve,
  title={EVOLvE: Evaluating and Optimizing LLMs For Exploration},
  author={Nie, Allen and Su, Yi and Chang, Bo and Lee, Jonathan N and Chi, Ed H and Le, Quoc V and Chen, Minmin},
  journal={arXiv preprint arXiv:2410.06238},
  year={2024}
}

@article{monea2024llms,
  title={LLMs Are In-Context Reinforcement Learners},
  author={Monea, Giovanni and Bosselut, Antoine and Brantley, Kiant{\'e} and Artzi, Yoav},
  journal={COLM},
  year={2025}
}

@article{park2024llm,
  title={Do llm agents have regret? a case study in online learning and games},
  author={Park, Chanwoo and Liu, Xiangyu and Ozdaglar, Asuman and Zhang, Kaiqing},
  journal={arXiv preprint arXiv:2403.16843},
  year={2024}
}

@article{coda2023meta,
  title={Meta-in-context learning in large language models},
  author={Coda-Forno, Julian and Binz, Marcel and Akata, Zeynep and Botvinick, Matt and Wang, Jane and Schulz, Eric},
  journal={Advances in Neural Information Processing Systems},
  volume={36},
  pages={65189--65201},
  year={2023}
}

@article{brown2020language,
  title={Language models are few-shot learners},
  author={Brown, Tom and Mann, Benjamin and Ryder, Nick and Subbiah, Melanie and Kaplan, Jared D and Dhariwal, Prafulla and Neelakantan, Arvind and Shyam, Pranav and Sastry, Girish and Askell, Amanda and others},
  journal={Advances in neural information processing systems},
  volume={33},
  pages={1877--1901},
  year={2020}
}

@article{lee2024supervised,
  title={Supervised pretraining can learn in-context reinforcement learning},
  author={Lee, Jonathan and Xie, Annie and Pacchiano, Aldo and Chandak, Yash and Finn, Chelsea and Nachum, Ofir and Brunskill, Emma},
  journal={Advances in Neural Information Processing Systems},
  volume={36},
  year={2024}
}

@article{yan2024efficient,
  title={Efficient Reinforcement Learning with Large Language Model Priors},
  author={Yan, Xue and Song, Yan and Feng, Xidong and Yang, Mengyue and Zhang, Haifeng and Ammar, Haitham Bou and Wang, Jun},
  journal={arXiv preprint arXiv:2410.07927},
  year={2024}
}

@inproceedings{carta2023grounding,
  title={Grounding large language models in interactive environments with online reinforcement learning},
  author={Carta, Thomas and Romac, Cl{\'e}ment and Wolf, Thomas and Lamprier, Sylvain and Sigaud, Olivier and Oudeyer, Pierre-Yves},
  booktitle={International Conference on Machine Learning},
  pages={3676--3713},
  year={2023},
  organization={PMLR}
}

@article{snell2024scaling,
  title={Scaling llm test-time compute optimally can be more effective than scaling model parameters},
  author={Snell, Charlie and Lee, Jaehoon and Xu, Kelvin and Kumar, Aviral},
  journal={arXiv preprint arXiv:2408.03314},
  year={2024}
}

@article{yao2024tree,
  title={Tree of thoughts: Deliberate problem solving with large language models},
  author={Yao, Shunyu and Yu, Dian and Zhao, Jeffrey and Shafran, Izhak and Griffiths, Tom and Cao, Yuan and Narasimhan, Karthik},
  journal={Advances in Neural Information Processing Systems},
  volume={36},
  year={2024}
}

@article{khalifa2023exploring,
  title={Exploring demonstration ensembling for in-context learning},
  author={Khalifa, Muhammad and Logeswaran, Lajanugen and Lee, Moontae and Lee, Honglak and Wang, Lu},
  journal={arXiv preprint arXiv:2308.08780},
  year={2023}
}

@article{zhang2022automatic,
  title={Automatic chain of thought prompting in large language models},
  author={Zhang, Zhuosheng and Zhang, Aston and Li, Mu and Smola, Alex},
  journal={arXiv preprint arXiv:2210.03493},
  year={2022}
}

@article{xiong2023dq,
  title={Dq-lore: Dual queries with low rank approximation re-ranking for in-context learning},
  author={Xiong, Jing and Li, Zixuan and Zheng, Chuanyang and Guo, Zhijiang and Yin, Yichun and Xie, Enze and Yang, Zhicheng and Cao, Qingxing and Wang, Haiming and Han, Xiongwei and others},
  journal={arXiv preprint arXiv:2310.02954},
  year={2023}
}

@article{tonglet2023seer,
  title={SEER: A Knapsack approach to Exemplar Selection for In-Context HybridQA},
  author={Tonglet, Jonathan and Reusens, Manon and Borchert, Philipp and Baesens, Bart},
  journal={arXiv preprint arXiv:2310.06675},
  year={2023}
}

@article{xia2024beyond,
  title={Beyond numeric awards: In-context dueling bandits with llm agents},
  author={Xia, Fanzeng and Liu, Hao and Yue, Yisong and Li, Tongxin},
  journal={arXiv preprint arXiv:2407.01887},
  year={2024}
}

@book{lattimore2020bandit,
  title={Bandit algorithms},
  author={Lattimore, Tor and Szepesv{\'a}ri, Csaba},
  year={2020},
  publisher={Cambridge University Press}
}

@article{reimers2019sentence,
  title={Sentence-BERT: Sentence Embeddings using Siamese BERT-Networks},
  author={Reimers, N},
  journal={arXiv preprint arXiv:1908.10084},
  year={2019}
}

@article{cer2018universal,
  title={Universal sentence encoder},
  author={Cer, D},
  journal={arXiv preprint arXiv:1803.11175},
  year={2018}
}

@misc{arxiv_api,
  author       = {{arXiv.org}},
  title        = {arXiv API},
  year         = {2025},
  url          = {https://arxiv.org/help/api},
  note         = {Accessed: 2025-01-28}
}

@misc{openai2025introducing,
  title = {Introducing Operator},
  author = {{OpenAI}},
  year = {2025},
  url = {https://openai.com/index/introducing-operator/},
  note = {Accessed: 2025-01-29}
}

@article{mukherjee2024pretraining,
  title={Pretraining decision transformers with reward prediction for in-context multi-task structured bandit learning},
  author={Mukherjee, Subhojyoti and Hanna, Josiah P and Xie, Qiaomin and Nowak, Robert},
  journal={arXiv preprint arXiv:2406.05064},
  year={2024}
}

@article{harper2015movielens,
  title={The movielens datasets: History and context},
  author={Harper, F Maxwell and Konstan, Joseph A},
  journal={Acm transactions on interactive intelligent systems (tiis)},
  volume={5},
  number={4},
  pages={1--19},
  year={2015},
  publisher={Acm New York, NY, USA}
}

@article{harris2026context,
  title={In-Context Credit Assignment via the Core},
  author={Harris, Keegan and Prasad, Siddharth and Trockman, Asher},
  journal={arXiv preprint arXiv:2605.06920},
  year={2026}
}

@inproceedings{carbonell1998use,
  title={The use of MMR, diversity-based reranking for reordering documents and producing summaries},
  author={Carbonell, Jaime and Goldstein, Jade},
  booktitle={Proceedings of the 21st annual international ACM SIGIR conference on Research and development in information retrieval},
  pages={335--336},
  year={1998}
}
